\documentclass{article}
\usepackage[preprint]{colm2026_conference}

\usepackage{microtype}
\usepackage{hyperref}
\usepackage{url}
\usepackage{booktabs}

\usepackage{lineno}

\definecolor{darkblue}{rgb}{0, 0, 0.5}
\hypersetup{colorlinks=true, citecolor=darkblue, linkcolor=darkblue, urlcolor=darkblue}

\usepackage[T1]{fontenc}

\usepackage[utf8]{inputenc}

\usepackage{graphicx}    %
\usepackage{subcaption}  %
\usepackage{adjustbox}
\usepackage{calc}
\usepackage{tikz}
\usepackage{multirow}
\usepackage{float}
\usepackage{amssymb}
\usepackage{wrapfig}
\usepackage{placeins} %
\usepackage{xspace}

\usepackage{amsmath,amsfonts,bm}

\def\eqref#1{equation~\ref{#1}}

\def\1{\bm{1}}

\DeclareMathAlphabet{\mathsfit}{\encodingdefault}{\sfdefault}{m}{sl}
\SetMathAlphabet{\mathsfit}{bold}{\encodingdefault}{\sfdefault}{bx}{n}

\newcommand{\aautoref}[1]{\hyperref[#1]{Appendix~\ref*{#1}}}

\newcommand{\eat}[1]{\ignorespaces}
\newcommand{\xxcomment}[4]{\textcolor{#1}{[$^{\textsc{#2}}_{\textsc{#3}}$ #4]}}

\newcommand{\citeit}[1]{\xxcomment{blue}{Y}{F}{citeit}}
\newcommand{\refit}[1]{\xxcomment{brown}{Y}{F}{refit}}

\newcommand{\evalname}{BenchPress\xspace}

\title{No Mean Feat: Simple, Strong Baselines \\ for Context \mbox{Compression}}

\author{Yair Feldman and Yoav Artzi \\
Department of Computer Science and Cornell Tech\\
Cornell University\\
\texttt{\{yairf,yoav\}@cs.cornell.edu}
}

\begin{document}

\ifcolmsubmission
\linenumbers
\fi

\maketitle

\begin{abstract}
Context compression reduces Transformer inference costs by replacing lengthy inputs with shorter pre-computed representations. It carries significant benefits for retrieval-augmented generation (RAG) and has attracted growing research attention. However, progress remains difficult to measure due to inconsistent evaluations and baselines. We design a standard, easy-to-reproduce evaluation suite for context compression, \emph{\evalname}, along with simple, high-performance baselines for English reading comprehension. \evalname supports benchmarking across model scales, datasets, compression ratios, and short ($<$1K tokens) to mid-range ($<$8K tokens) contexts. While the suite is applicable to any compression paradigm, our baselines target soft context compression.
We establish two simple baselines that strongly outperform the widely used causal compression-token approach: mean pooling and a bidirectional compression-token variant. Our results show the benefit of bidirectional attention when computing compressed representations, and that simple pooling is an expressive compression operator.

\end{abstract}

\section{Introduction}

Repeated reasoning over long documents is core for retrieval-augmented generation (RAG), where the same evidence is likely repeatedly processed across queries. This is computationally costly, both in time (processing and attending over the document) and memory (the key-value cache). \emph{Context compression} addresses these costs through strategies such as soft compression, key-value (KV) cache compression, and hard prompt compression~\citep[e.g.,][]{ge2024incontext, xrag, dai-etal-2025-pretraining}.

Despite growing interest, progress remains difficult to measure. Evaluations differ in datasets, metrics, context lengths, and model scales, and even within soft compression the most common baseline, causal compression tokens, sets a low bar. Without a shared evaluation framework, it is unclear whether reported improvements represent real advances or favorable experimental choices. We address this gap by making two contributions. 

First, we design a standardized, easy-to-reproduce evaluation suite for context compression, \emph{\evalname} (\autoref{sec:eval}). Although applicable to any compression paradigm, we instantiate it here for soft context compression, where we find that even simple methods can meaningfully advance the state of the art. The suite covers short ($<$1K tokens) and mid-range ($<$8K tokens) contexts with an explicit in-domain/out-of-domain split, teacher-normalized scoring for fair cross-model comparison, and a standardized training mixture to disentangle methodological contributions from data effects.

Second, we establish two simple but strong baseline methods for soft context compression. The first, \emph{mean pooling}, is a compression operator that averages adjacent hidden states produced by a bidirectionally-encoding fine-tuned LLM, introducing no parameters beyond the encoder backbone. The second, \emph{bidirectional compression tokens}, is a simple modification to the widely-used causal compression-token approach in which compression tokens attend bidirectionally among themselves, making the model aware of its compression budget, while retaining causal attention over the context. This modification has not been explored in prior work, despite its simplicity. Both baselines substantially outperform the standard causal compression-token approach.

Using our evaluation suite, we find that mean pooling achieves the strongest results overall, particularly at ratios up to 16$\times$; at 128$\times$, bidirectional tokens are competitive or superior in several settings. A shared insight is that \emph{bidirectional attention during encoding is critical for compression quality}. We also find that multi-ratio training is feasible with minor degradation for mean pooling, while bidirectional tokens even benefit from it. Compression quality scales favorably with model size, and mean pooling's advantages are even larger at longer contexts (8K tokens).
We release our code and data at \url{https://github.com/lil-lab/benchpress}.

\section{Task Definition}

we formally define \emph{soft} context compression, the paradigm targeted by our baselines, although  our evaluation suite (\autoref{sec:eval}) is applicable to any compression paradigm. In the soft compression setting, a document of length $L$ is mapped to a sequence of dense continuous vectors of length $C$, with $L \gg C$.
This process allows an LLM that uses the compressed version of the document to invest significantly less computation, both in time and KV cache space, both reduced from dependence on $L$ to dependence on $C$.
This benefit increases with repeated use of the document, as is likely in RAG scenarios.

We define soft context compression to support flexible compression ratios.
Let $\mathcal{M}$ be a language model and $\mathcal{R} \subseteq \mathbb{N_+}$  the admissible set of compression ratios. 
Let $\mathcal{V}$ denote the vocabulary and $d$ the embedding dimension of $\mathcal{M}$. 
The goal of learning is to construct a compression function
\begin{equation}
f_c : \mathcal{V}^{L} \times \mathcal{R} \to \mathbb{R}^{C \times d}\;\;,
\end{equation}
which maps a token sequence $T = (t_{1}, \ldots, t_{L})$, $t_i \in \mathcal{V}$ of length $L$ and a ratio $r \in \mathcal{R}$ to a compressed representation of $C$ vectors of dimension $d$. The length $C$ is determined by the specified ratio $C = \left\lceil L / r \right\rceil$.

An ideal compressor $f_c$ preserves the conditional distribution of the model using the compressed version for any prompt $P$:
\begin{equation}
p_{\mathcal{M}}(\cdot \mid T, P) \;\approx\; p_{\tilde{\mathcal{M}}}(\cdot \mid f_c(T; r), P)\;\;,
\end{equation}
where $\tilde{\mathcal{M}}$ is a potentially adapted version of $\mathcal{M}$, for example augmented with lightweight parameter fine-tuning via LoRA~\citep{hu2022lora} modules that can be fused into the base model without altering its capacity.

\section{Background and Related Work}\label{sec:background}

\begin{figure*}[t]
    \centering
    
    \def\uniformscale{0.23} %
    
    \newlength{\fixedheight}
    \setlength{\fixedheight}{5.3cm}
    
    \begin{subfigure}[t]{0.31\textwidth}
        \centering
        \vbox to \fixedheight{%
            \includegraphics[page=1, scale=\uniformscale]{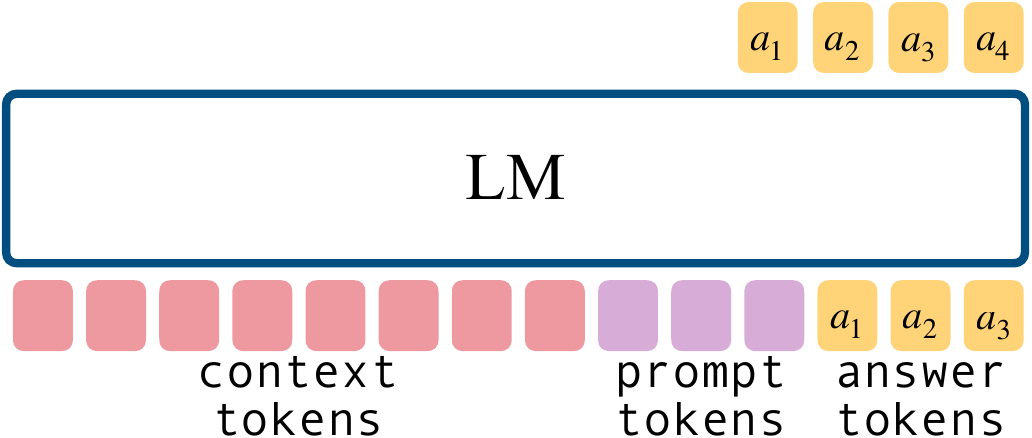}%
            \vfill
        }
        \caption{Processing with a regular language model (no compression).}
        \label{fig:arch:regular_llm}
    \end{subfigure}
    \hfill
    \begin{minipage}[t]{0.02\textwidth}
        \centering
        \rule[0pt]{0pt}{\fixedheight}%
        \begin{tikzpicture}
            \draw[dashed, gray, very thin] (0,0) -- (0,\fixedheight);
        \end{tikzpicture}
    \end{minipage}%
    \hfill
    \begin{subfigure}[t]{0.31\textwidth}
        \centering
        \vbox to \fixedheight{%
            \includegraphics[page=2, scale=\uniformscale]{figures/architecture_fig.pdf}%
            \vfill
        }
        \caption{Compression tokens approach for context compression.}
        \label{fig:arch:tokens}
    \end{subfigure}
    \hfill
    \begin{minipage}[t]{0.02\textwidth}
        \centering
        \rule[0pt]{0pt}{\fixedheight}%
        \begin{tikzpicture}
            \draw[dashed, gray, very thin] (0,0) -- (0,\fixedheight);
        \end{tikzpicture}
    \end{minipage}%
    \hfill
    \begin{subfigure}[t]{0.31\textwidth}
        \centering
        \vbox to \fixedheight{%
            \includegraphics[page=3, scale=\uniformscale]{figures/architecture_fig.pdf}%
            \vfill
        }
        \caption{Mean-pooling baseline: no extra tokens; mean pooling of final hidden states.}
        \label{fig:arch:pool}
    \end{subfigure}

    \caption{Context processing strategies compared in our benchmark: (a) regular LM with full context, (b) compression tokens, and (c) our mean pooling baseline. The figure illustrates a compression ratio of $4\times$.}
    \label{fig:arch}
\end{figure*}

\paragraph{Soft Context Compression}
A dominant line of research on context compression uses \emph{compression tokens}.
As shown in \autoref{fig:arch:tokens}, a sequence of length $L$ is augmented with $C = \lceil L/r \rceil$ additional tokens.\footnote{Some approaches use a fixed number of compression tokens with distinct learned embeddings per position.}
The final hidden states at the compression-token positions form the compressed representation, which a decoder then conditions on alongside a downstream prompt.
Training typically combines a language modeling objective with a distillation loss that encourages the compressor--decoder to approximate a teacher LLM with full context access (\autoref{fig:arch:regular_llm}).

This problem is getting substantial research attention. 
We list here representative examples. 
AutoCompressors~\citep{chevalier2023adapting} introduce recursive compression with tied encoder--decoder weights.
ICAE~\citep{ge2024incontext} freezes the decoder and trains only the encoder via autoencoding pretraining followed by task finetuning.
COCOM~\citep{cocom} extends this to retrieval-augmented QA with lighter encoders and joint multi-context decoders.
xRAG~\citep{xrag} maps retrieval embeddings directly to the decoder’s input space, achieving single-token compression but with constraints on generality.
PISCO~\citep{louis-etal-2025-pisco} trains compressors on LLM-generated answers to improve RAG performance, while PCC~\citep{dai-etal-2025-pretraining} learns a converter to project compressed representations across model boundaries.
GMSA~\citep{tang2025gmsaenhancingcontextcompression} groups hidden representations with a layer semantic alignment module, which is related to our pooling study but relies on multi-stage reconstruction training and compression--decoder adapters.\footnote{We were unable to include GMSA in our evaluation as their code and models were not publicly available at the time of writing.}

Comparisons across these methods remain difficult due to inconsistent evaluation setups, metrics, and baselines, a gap we address with a standardized evaluation suite and simple, strong baselines.

\paragraph{KV Cache Compression}
In contrast to representing contexts as input embeddings, another line of work compresses the entire set of key–value (KV) states. 
Some approaches remove or compress less informative entries in the KV cache without additional training~\citep{xiao2024efficient_streamingllm, oren-etal-2024-transformers_TOVA,li2024snapkv}, while others train the model to perform the compression explicitly~\citep{qin-etal-2024-dodo,DynamicMemoryCompression}. 
A different variant introduces compression tokens, but instead of retaining only the final hidden representation, all KV states are propagated to the decoder~\citep[e.g.,][]{zhang2025long_activationbeacon,li-etal-2025-500xcompressor}. 
Although these methods provide higher-capacity compressed representations that are well suited for efficient long-context comprehension, their increased size of the compressed representations raises complex practical storage and networking challenges for RAG frameworks, where caching compressed representations could otherwise avoid recomputation. 

\paragraph{Hard Prompt Compression}
An alternative approach is to compress contexts directly in the token space. 
This has been done by removing unimportant tokens or lexical units~\citep[e.g.,][]{li-etal-2023-selectivecontext,jiang-etal-2023-llmlingua, pan-etal-2024-llmlingua2} or generating concise summaries that preserve salient details~\citep{chuang-etal-2024-learning_NanoCapsulator}. 
While these methods can be more interpretable and storage-efficient, they are inherently constrained by their reliance on explicit tokens.

\section{The \evalname Evaluation Suite}\label{sec:eval}

Our evaluation suite is designed around components that are not specific to any single compression paradigm: its datasets, metrics, and protocol can be applied to soft compression, KV cache methods, and hard prompt compression alike. We demonstrate this breadth by evaluating both soft compression baselines and a hard-prompt method (LLMLingua2) within the same framework. A central design principle is to isolate compression quality from retrieval noise: we focus on reading comprehension, where contexts are guaranteed to contain the necessary evidence, enabling controlled comparison across datasets that stress both single-hop and multi-hop reasoning.

\subsection{Datasets}\label{sec:eval:data}

The suite comprises two evaluation tiers spanning different context lengths, together covering short ($<$1K tokens) and mid-range ($<$8K tokens) inputs.

\paragraph{Short-context Benchmarks} Six reading comprehension datasets form the primary evaluation (\autoref{tab:evaluation_datasets}): SQuAD~\citep{rajpurkar-etal-2016-squad}, NarrativeQA~\citep{kocisky-etal-2018-narrativeqa}, HotpotQA~\citep{yang-etal-2018-hotpotqa}, AdversarialQA~\citep{bartolo2020beat_adversarialqa}, TriviaQA~\citep{joshi-etal-2017-triviaqa} (verified subset), and ParaphraseRC~\citep{saha-etal-2018-duorc}. This selection spans reasoning styles from factual extraction to adversarial paraphrasing. The benchmark explicitly separates in-domain and out-of-domain evaluation: training includes SQuAD, NarrativeQA, and HotpotQA (in-domain), while AdversarialQA, TriviaQA, and ParaphraseRC are held out entirely (out-of-domain).\footnote{In- vs.\ out-of-domain results are in \autoref{app:in_vs_out}.}

\begin{table}[t]
\centering
\begin{tabular}{lrrr}
\toprule
Dataset & Avg. Context Tokens & \#Samples & \#Contexts \\
\midrule
AdversarialQA~\citep{bartolo2020beat_adversarialqa}      & 154 & 1{,}000 & 341 \\
HotpotQA~\citep{yang-etal-2018-hotpotqa}           & 254 & 7{,}394 & 7{,}352 \\
NarrativeQA~\citep{kocisky-etal-2018-narrativeqa}        & 639 & 3{,}002 & 100 \\
ParaphraseRC~\citep{saha-etal-2018-duorc}       & 685 & 4{,}835 & 560 \\
SQuAD~\citep{rajpurkar-etal-2016-squad}              & 169 & 5{,}928 & 1{,}204 \\
TriviaQA (Verified)~\citep{joshi-etal-2017-triviaqa} & 539 & 185   & 185 \\
\midrule
\textbf{Total}     & \textbf{375} & \textbf{22{,}344} & \textbf{9{,}742} \\
\bottomrule
\end{tabular}
\caption{Short-context evaluation datasets. The overall average context length is weighted by the number of samples.}
\label{tab:evaluation_datasets}
\end{table}

\paragraph{Mid-range Context Benchmarks} To assess how compression methods scale to longer inputs, where computational savings are more valuable, we include QA tasks from LongBench-E~\citep{bai2024longbench} with contexts up to 8K tokens (\autoref{tab:longbench-datasets}).

\begin{table}[t]
\centering
\begin{tabular}{lrrr}
\toprule
Dataset & Avg. Context Tokens & \#Samples & \#Contexts \\
\midrule
\multicolumn{4}{l}{\textbf{Single-Doc QA}} \\
QASPER~\citep{dasigi-etal-2021-qasper}      & 4{,}901 & 192 & 133 \\
MultiFieldQA-en~\citep{bai2024longbench}           & 4{,}725 & 93 & 66 \\
\addlinespace
\multicolumn{4}{l}{\textbf{Multi-Doc QA}} \\
HotpotQA~\citep{yang-etal-2018-hotpotqa}        & 4{,}997 & 128 & 128 \\
2WikiMultihopQA~\citep{xanh2020_2wikimultihop}       & 5{,}426 & 165 & 165 \\
\midrule
\textbf{Total}     & \textbf{5{,}044} & \textbf{578} & \textbf{492} \\
\bottomrule
\end{tabular}
\caption{Mid-range context length LongBench-E evaluation datasets. The overall average context length is weighted by the number of samples. We only include samples for which the context length is under 8K tokens.}
\label{tab:longbench-datasets}
\end{table}

\subsection{Training Data}\label{sec:eval:training}

A key component of the evaluation suite is a standardized training mixture. Performance differences between compression methods can stem from the method itself or from the training data; without controlling for both, it is impossible to attribute gains cleanly. Our training mixture draws from the train splits of the in-domain QA datasets as well as summarization tasks; full details appear in \autoref{tab:training_datasets}. As in the construction of the evaluation suite, our guiding principle is to only include datasets whose contexts are guaranteed to contain the necessary evidence for completing the task at hand. We recommend that future evaluations use a shared training mixture, or at minimum report the mixture used, so that methodological contributions can be disentangled from data effects.

\subsection{Metrics}\label{sec:eval:metrics}

We evaluate using \emph{exact match} (EM) and $F_1$.\footnote{We show only $F_1$ in the main text; \autoref{tab:primary_results_EM} shows full EM results.} We do not use the substring accuracy metric (score of 1 if the gold answer is a substring of the model output) as it is easily exploitable,\footnote{E.g., listing all 50 US states as the answer to any state-valued question.} which forces us to exclude some baselines from primary comparisons.

For each metric we define a \emph{teacher-normalized} version. Given metric $M$, teacher $\mathcal{M}$, and compressor $f_c$: let $M_T$ be the teacher's score with full context, $M^{\varnothing}_T$ the no-context score, and $M_{f_c}$ the score with compressed context. The teacher-normalized score is:
\[
\frac{M_{f_c} - M^{\varnothing}_T}{M_T - M^{\varnothing}_T}\;\;.
\]
This scales performance relative to the teacher and corrects for questions answerable without context, enabling fair cross-model comparison regardless of teacher quality.

\subsection{Reference Systems}\label{sec:eval:systems}

Although the suite can accommodate any compression paradigm, in this work we focus on soft context compression, where we find that simple baselines can improve markedly over existing practice. In addition to the two baselines we introduce (\autoref{sec:baselines}), we evaluate several existing soft compression methods: ICAE~\citep{ge2024incontext} and PCC~\citep{dai-etal-2025-pretraining}. To demonstrate the suite's cross-paradigm applicability, we also evaluate LLMLingua2~\citep{pan-etal-2024-llmlingua2}, a hard-prompt compression approach, by passing its compressed prompts to our finetuned Qwen3-8B teacher model.

Comparison across methods is challenging due to inconsistencies in training procedures, available code, and supported metrics. Our primary goal is to map the architecture landscape systematically. For example, PISCO~\citep{louis-etal-2025-pisco} was evaluated only using substring accuracy; we omit it from primary comparisons but report substring accuracy in \aautoref{app:results:metrics} (\autoref{tab:primary_results_accuracy}). These difficulties further motivate standardized evaluation.

\section{Baseline Methods}\label{sec:baselines}

\newcommand{\context}{T}
\newcommand{\prompt}{P}
\newcommand{\numpoolops}{b}
\newcommand{\lencompression}{C}
\newcommand{\lencontext}{L}
\newcommand{\comprate}{r}
\newcommand{\allcomprates}{\mathcal{R}}
\newcommand{\embdim}{d}
\newcommand{\encodingseq}{h}
\newcommand{\encoding}{h}

We establish two simple baselines for soft context compression, both trained via knowledge distillation from a teacher LLM with access to the full uncompressed context. The first baseline, \emph{mean pooling}, is a compression operator that averages adjacent hidden states after bidirectional encoding, without adding additional parameters beyond the encoder. The second, \emph{bidirectional compression tokens}, is a straightforward modification to the widely-used causal compression-token approach in which the compression tokens attend bidirectionally among themselves. Both baselines are considerably stronger than the standard causal compression-token approach. 
\autoref{fig:arch} provides an overview of the processing strategies we compare.

\subsection{Baseline 1: Mean Pooling}\label{sec:baselines:pool}

The first baseline compresses context representations via non-overlapping mean pooling over a bidirectionally-encoded sequence.
\autoref{fig:arch:pool} illustrates this approach.
Given a document, we compute its representation with an encoder, and apply a non-overlapping mean pooling operator with window size $\comprate$, the same size as the compression ratio, and stride $\comprate$ to generate continuous vectors as the output compression.

Formally, let $\encodingseq = (\encoding_1,\ldots,\encoding_\lencontext) \in \mathbb{R}^{\lencontext \times \embdim}$ denote the sequence of hidden states produced by the encoder.
For a compression ratio $\comprate \in \allcomprates$, we partition the sequence into $k$ consecutive, non-overlapping blocks:
\begin{equation}
\begin{gathered}
    S_k = \{(k-1)\comprate+1,\ldots, \quad \min(k\comprate, \lencontext)\}, \\
    k = 1,\ldots,\lceil \lencontext/\comprate \rceil.
\end{gathered}
\end{equation}
The compressed representation of length $\lceil \lencontext/\comprate \rceil$ is obtained by averaging within each block:
\begin{equation}
\begin{gathered}
f_c(T,\comprate) = (z_1,\ldots,z_{\lceil \lencontext/\comprate \rceil}) \in \mathbb{R}^{\lceil \lencontext/\comprate \rceil \times d} \hspace{1em}, \\
\text{s.t.}\hspace{1em}z_k = \frac{1}{|S_k|} \sum_{i \in S_k} h_i\;\;.
\end{gathered}
\end{equation}

The encoder is Transformer-based. Critically, we use a full self-attention mask during encoding, allowing each encoded vector to include information from the entire context, and thereby each compressed segment to aggregate information across the entire context. In practice, we initialize with an autoregressive LLM and remove the causal self-attention mask before learning.

This design introduces no additional parameters beyond those of the encoder backbone and the decoder LLM that uses the compressed representation, and has negligible computational overhead. By comparison, the compression-token method is slightly more expensive: it requires an encoder input of size $\lencontext + \lencontext/\comprate$, while mean pooling processes only the original $\lencontext$ tokens.

\subsection{Baseline 2: Bidirectional Compression Tokens}\label{sec:baselines:tokens}

The widely-used causal compression-token approach (described in \autoref{sec:background}) appends $C = \lceil L/r \rceil$ compression tokens to a context of length $L$ and extracts their final hidden states as the compressed representation. 
This choice has an interesting implication, especially if aiming to support multiple compression ratios, as we additionally study in our experiments. 
The standard causal attention mask imposes a Matryoshka-style~\citep{kusupati2022matryoshka} constraint: compressed representations at smaller lengths must correspond to strict prefixes of those at larger lengths.

Our second baseline relaxes this restriction by allowing compression tokens to attend bidirectionally among themselves, while retaining causal attention over the original context. This modification, not explored in prior work despite its simplicity, makes the model aware of its compression budget, enabling ratio-specific information allocation while preserving shared computation and KV caching. We experiment with both the conventional causal mask and this bidirectional variant; the modification substantially improves performance (\autoref{sec:exp:results}).

In our implementation, we utilize only a single compression token type that is appended $\lceil \lencontext/\comprate \rceil$ times to the context, rather than having several distinct compression tokens. This enables compression to any arbitrary ratio while retaining equal parameter counts across ratio settings.

\subsection{Training Objective}\label{sec:baselines:train}

\newcommand{\answertoken}{a}
\newcommand{\answerseq}{A}

Both baselines are trained via knowledge distillation. Each training instance comprises a context $\context$, a prompt $\prompt$, and a ground-truth answer $\answerseq = (\answertoken_1,\ldots,\answertoken_m)$. At each step $i$, let $Q_i = q(\cdot \mid \context, \prompt, \answerseq_{<i})$ denote the teacher's distribution given the full context. Similarly, let $P_{\theta,i} = p_\theta(\cdot \mid \tilde{\context}, \prompt, \answerseq_{<i})$ be the student's distribution, where $\tilde{\context} = f_c(\context, \comprate)$ represents the compressed context at ratio $\comprate$.

The distillation loss is the sum of the KL divergences between these step-wise distributions:
\begin{equation}
    \mathcal{L}_{\text{KD}}(\context, \prompt, \answerseq; \comprate) =
    \sum_{i=1}^m \mathrm{KL}\big( Q_i \;\|\; P_{\theta,i} \big).
\end{equation}

We also study a \emph{multi-ratio training} regime where a single compressor handles multiple compression ratios simultaneously. For each training instance, we generate compressed representations for all $r \in \mathcal{R}$, pass each independently to the decoder, and sum the losses:
\begin{equation}
\mathcal{L}_{\text{multi}}(\context,\prompt,\answerseq) \;=\;
\sum_{\comprate \in \allcomprates}
\mathcal{L}_{\text{KD}}(\context,\prompt,\answerseq; \comprate)\;\;.
\end{equation}
A single parameter update is applied after aggregating losses. Since encoder computation is shared across ratios, this is considerably more efficient than training separate models. Using only knowledge distillation, without a language modeling objective, enables direct comparison between the original model and each compressor.

\subsection{Model Training}\label{sec:baselines:training}

For each target LLM, we first finetune it on the training mixture using LoRA~\citep{hu2022lora}; this finetuned model is fixed as the teacher, ensuring performance differences stem from the compressor alone. Both encoder and decoder are initialized from the same LLM but trained with separate LoRA weights, always using instruction-tuned weights as the backbone. For multi-ratio training, we train on $\{4\times, 8\times, 16\times, 32\times, 64\times, 128\times\}$ unless stated otherwise. A learned linear projection $W\in\mathbb{R}^{d\times d}$ is applied to $f_c(T,\comprate)$ before passing to the decoder, as it slightly improves performance for both baselines.

\paragraph{Long-Context Training}
Our primary experiments use contexts up to 1K tokens. For longer inputs (up to 8K), we use a separate data mixture and a staged training procedure that extends context length while preserving short-context performance, using Qwen3-1.7B (pretrained at 32K). Full details are in \aautoref{app:exp:long}.

\section{Experiments and Results}\label{sec:exp}

We evaluate the baselines from \autoref{sec:baselines} using the suite from \autoref{sec:eval}, comparing causal compression tokens, bidirectional compression tokens, and mean pooling. To demonstrate the suite's generality, we run experiments across six models spanning three families and four scales: Llama3.2-1B~\citep{grattafiori2024llama}, Gemma2-2B~\citep{team2024gemma}, and Qwen3-0.6/1.7/4/8B~\citep{yang2025qwen3}.

\subsection{Short-Context Results}\label{sec:exp:results}

\begin{table*}[!ht]
    \small
    \centering
    \setlength{\tabcolsep}{5pt}
    \resizebox{\textwidth}{!}{%
\begin{tabular}{llcccccccc}
\toprule
\multicolumn{2}{c}{} & \multicolumn{1}{c}{\textbf{Original}} & \multicolumn2{c}{\textbf{4x}} & \multicolumn2{c}{\textbf{16x}} & \multicolumn2{c}{\textbf{128x}} & \multicolumn{1}{c}{\textbf{No Ctx}} \\
\cmidrule(lr){3-3}\cmidrule(lr){4-5}\cmidrule(lr){6-7}\cmidrule(lr){8-9}\cmidrule(l){10-10}
\multicolumn{2}{c}{} &  & \textbf{Single} & \textbf{Multi} & \textbf{Single} & \textbf{Multi} & \textbf{Single} & \textbf{Multi} &  \\
\midrule
\multicolumn{10}{l}{\textbf{Baseline Systems}} \\
& \emph{LLMLingua2} (Qwen3-8B) &  & & 42.52 &  & 24.39 &  & 22.59 &  \\
& \emph{ICAE} (Mistral-7B) &  & 42.40 &  &  &  &  &  &  \\
& \emph{PCC Lite} (GPT2-Large \& Llama3.1-8B) &  & 62.08 &  & 51.30 &  & 36.20 &  &  \\
& \emph{PCC Large} (Llama3.1-8B) &  & 62.98 &  & 49.37 &  & 37.24 &  &  \\
\midrule
\multicolumn{10}{l}{\textbf{Our Baselines}} \\
\addlinespace[2pt]
\multicolumn{2}{l}{\textbf{Qwen3-8B}} & 74.33 &  &  &  &  &  &  & 23.06 \\
 & Compression-Tokens (Causal) &  & 67.03 & 65.90 & 56.21 & 58.41 & 47.47 & 44.76 &  \\
 & Compression-Tokens (Bidirectional) &  & 69.20 & 69.57 & 60.27 & 63.01 & 46.93 & \textbf{46.97} &  \\
 & Mean-Pooling &  & \textbf{71.66} & \textbf{70.55} & \textbf{63.85} & \textbf{64.67} & \textbf{47.90} & 45.92 &  \\
\midrule
\multicolumn{2}{l}{\textbf{Qwen3-4B}} & 73.44 &  &  &  &  &  &  & 19.79 \\
 & Compression-Tokens (Causal) &  & 64.88 & 62.53 & 55.22 & 54.28 & 43.08 & 40.83 &  \\
 & Compression-Tokens (Bidirectional) &  & 66.72 & 68.15 & 57.68 & 60.48 & 41.61 & \textbf{42.66} &  \\
 & Mean-Pooling &  & \textbf{70.39} & \textbf{69.36} & \textbf{61.79} & \textbf{61.72} & \textbf{43.62} & 41.05 &  \\
\midrule
\multicolumn{2}{l}{\textbf{Qwen3-1.7B}} & 69.93 &  &  &  &  &  &  & 14.00 \\
 & Compression-Tokens (Causal) &  & 50.90 & 57.73 & 49.83 & 48.68 & 36.19 & 35.34 &  \\
 & Compression-Tokens (Bidirectional) &  & 62.04 & 62.60 & 51.53 & 54.11 & 36.25 & \textbf{35.77} &  \\
 & Mean-Pooling &  & \textbf{66.43} & \textbf{64.17} & \textbf{55.43} & \textbf{54.47} & \textbf{36.72} & 33.48 &  \\
\midrule
\multicolumn{2}{l}{\textbf{Qwen3-0.6B}} & 65.36 &  &  &  &  &  &  & 9.34 \\
 & Compression-Tokens (Causal) &  & 54.40 & 51.85 & 41.57 & 42.59 & 28.86 & 28.60 &  \\
 & Compression-Tokens (Bidirectional) &  & 55.59 & 57.03 & 44.82 & 47.62 & 29.69 & \textbf{29.51} &  \\
 & Mean-Pooling &  & \textbf{61.17} & \textbf{58.36} & \textbf{47.59} & \textbf{47.64} & \textbf{29.94} & 26.36 &  \\
\midrule
\multicolumn{2}{l}{\textbf{Gemma2-2B}} & 71.96 &  &  &  &  &  &  & 21.64 \\
 & Compression-Tokens (Causal) &  & 63.35 & 62.18 & 55.07 & 54.70 & 44.46 & 42.49 &  \\
 & Compression-Tokens (Bidirectional) &  & 64.76 & 65.24 & 56.39 & 58.43 & 44.73 & 43.17 &  \\
 & Mean-Pooling &  & \textbf{69.33} & \textbf{68.09} & \textbf{61.39} & \textbf{61.04} & \textbf{44.98} & \textbf{43.71} &  \\
\midrule
\multicolumn{2}{l}{\textbf{Llama3.2-1B}} & 65.82 &  &  &  &  &  &  & 15.17 \\
 & Compression-Tokens (Causal) &  & 56.31 & 53.74 & 47.51 & 46.96 & 35.41 & 35.62 &  \\
 & Compression-Tokens (Bidirectional) &  & 57.91 & 57.52 & \textbf{49.20} & 50.06 & \textbf{36.43} & \textbf{36.25} &  \\
 & Mean-Pooling &  & \textbf{62.81} & \textbf{60.56} & 47.28 & \textbf{51.56} & 33.25 & 33.98 &  \\
\midrule
\bottomrule
\end{tabular}}

\medskip
\includegraphics[width=\textwidth]{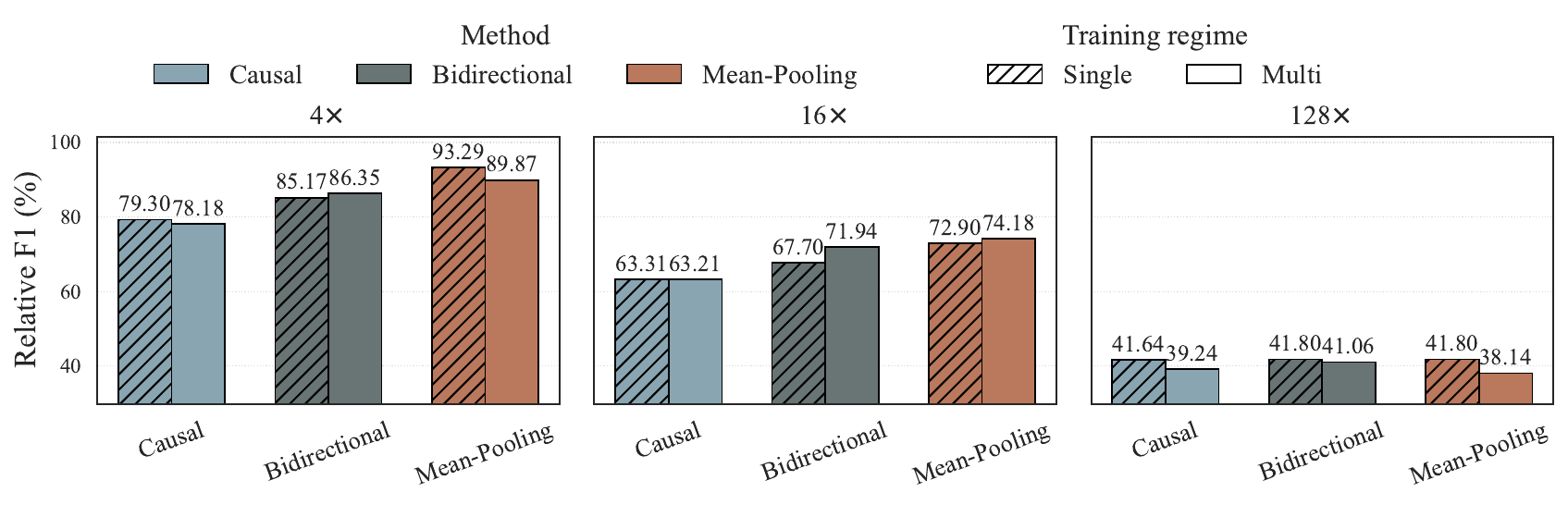}
\caption{
Primary results. Values in the table are $F_1$ scores macro-averaged across all datasets in our evaluation suite. 
\textbf{Original} stands for the teacher model's score when given the full context. \textbf{No Ctx} stands for the teacher model's score when not given any context at all. For each ratio, we display both single- and multi-ratio versions. For the baseline systems (top section), we include results for the compression ratios supported by these methods, unsupported ratios are left blank. The best method for each (model, ratio, single/multi-ratio training) setting is \textbf{bolded}.
Bottom figures present aggregated views of the results in the table, but instead of $F_1$ show the teacher-normalized $F_1$ metric (Relative F1). Scores are obtained by averaging over all models listed in the table. 
}\label{tab:primary_results_hybrid_transpose}
\end{table*}

\autoref{tab:primary_results_hybrid_transpose} presents benchmark results across all models and compression ratios. We show the average $F_1$ scores over all six evaluation datasets. Bar charts below the table summarize results as teacher-normalized $F_1$ averaged across all models and datasets, showing the teacher with full context (``Original''), without context (``No Ctx''), and the compressors at each ratio under single- and multi-ratio training.

Mean pooling achieves the strongest results overall, particularly at ratios up to 16$\times$. Adding bidirectional attention among compression tokens improves over the causal variant in most settings, with gains ranging from modest to over 11 F1 points depending on model and ratio. At extreme compression (128$\times$), bidirectional tokens are competitive with or superior to mean pooling in several settings; Llama3.2-1B consistently shows this pattern. Bidirectional compression tokens benefit markedly from multi-ratio training, while mean pooling shows a modest trade-off. A plausible explanation is that the bidirectional attention pattern lets compression tokens ``see'' how many tokens are available, giving the model a clear signal about its compression budget; mean pooling, by contrast, must produce a one-size-fits-all representation at each position.
\subsection{Long-Context Results}\label{sec:exp:long}
\begin{table}[!ht]
    \small
    \centering
    \setlength{\tabcolsep}{5pt}
\begin{tabular}{lcccccc}
\toprule
 & \multicolumn{3}{c}{\textbf{Single-Doc QA}} & \multicolumn{3}{c}{\textbf{Multi-Doc QA}} \\
\cmidrule(lr){2-4} \cmidrule(l){5-7}
 & \textbf{4x} & \textbf{16x} & \textbf{128x} & \textbf{4x} & \textbf{16x} & \textbf{128x} \\
\midrule
\multicolumn{7}{l}{\textbf{Baseline Systems}} \\
\textit{LLMLingua2} (Qwen3-1.7B) & 20.7 & 12.5 & 8.9 & 29.4 & 22.4 & 21.8 \\
\textit{PCC Large} (Llama3.1-8B) & 17.2 & 5.6 & 2.4 & 28.0 & 11.4 & 7.2 \\
\midrule
\multicolumn{7}{l}{\textbf{Our Baselines} (Qwen3-1.7B)} \\
Compression-Tokens (Causal) & 33.3 & 19.5 & 17.9 & 40.9 & 32.5 & 31.6 \\
Compression-Tokens (Bidirectional) & 35.9 & 30.5 & 19.1 & 43.0 & 38.2 & 32.1 \\
Mean-Pooling & \textbf{39.7} & \textbf{32.5} & \textbf{24.2} & \textbf{45.9} & \textbf{41.4} & \textbf{36.0} \\
\midrule
\textbf{Teacher Models} (Qwen3-1.7B) & \multicolumn{3}{c}{\textit{(no compression)}} & \multicolumn{3}{c}{\textit{(no compression)}} \\
\textit{w/o Context} & \multicolumn{3}{c}{\rule[.5ex]{2.5em}{0.3pt} 10.1 \rule[.5ex]{2.5em}{0.3pt}} & \multicolumn{3}{c}{\rule[.5ex]{2.5em}{0.3pt} 21.7 \rule[.5ex]{2.5em}{0.3pt}} \\
\textit{w/ Context} & \multicolumn{3}{c}{\rule[.5ex]{2.5em}{0.3pt} 43.3 \rule[.5ex]{2.5em}{0.3pt}} & \multicolumn{3}{c}{\rule[.5ex]{2.5em}{0.3pt} 49.6 \rule[.5ex]{2.5em}{0.3pt}} \\
\bottomrule
\end{tabular}
    \caption{LongBench-E QA results. Displayed metric is $F_1$. We include contexts with up to a maximum of 8K tokens. The best method for each ratio-dataset setting is \textbf{bolded}.}
    \label{tab:longbench_results}
\end{table}

We extend evaluation to the mid-range context tier (up to 8K tokens). \autoref{tab:longbench_results} shows QA results on LongBench-E: mean pooling remains superior, with margins over compression-token approaches that are even larger than in the short-context setting.

\subsection{Additional Findings}\label{sec:exp:additional}

\paragraph{Compression Scaling}
Compression quality scales favorably with model size: teacher-normalized $F_1$ increases across four Qwen3 scales (0.6B--8B), indicating that larger models retain more information under compression (\autoref{fig:scaling_law} in \aautoref{app:scaling}). Our experiments only consider compressors in which the encoder and decoder are of the same size. We leave the investigation of individual component scaling for future work. 

\paragraph{Ablations}
We ablate the mean-pooling baseline using Gemma2-2B (\autoref{tab:abl_gemma2_mean} in \aautoref{app:ablations}). The encoder is the most critical component ($>$12\% drop when frozen or removed), while the linear layer and ratio-sampling trade-offs are minor (0.7\% and 1.4\%, respectively).

\section{Discussion}\label{sec:discussion}

We create \evalname, an evaluation suite that provides a controlled, reproducible framework, not tied to any single compression paradigm, for measuring progress in this space. 
This is largely motivated by inconsistent evaluation practices in existing work, which entail comparison challenges, for example in not accounting for the capabilities of simple baselines, as our experiments show. 

We show that the causal compression-token approach is a weak baseline: mean pooling and bidirectional compression tokens both leverage bidirectional encoding to achieve considerably stronger results. This convergence suggests that causal attention is a poor inductive bias for compression, pointing toward encoder-style or prefix-LM backbones as more natural starting points. Interestingly, bidirectional tokens benefit from multi-ratio training while mean pooling generally does not, plausibly because forward attention gives the model an explicit view of its compression budget. Testing this hypothesis directly is a promising direction for future work.

A persistent gap between all methods and the teacher at 128$\times$ suggests that advances in compressor architecture are needed alongside scaling. We hope these standardized practices and strong baselines provide a foundation for future work across compression paradigms.

\section*{Acknowledgements}\label{sec:acknowledgements}

This research was supported by a gift to the LinkedIn–Cornell Bowers Strategic Partnership, NSF under grant No. OAC-2311521, NASA under award No. 20-OSTFL20-0053, and a gift from Open Philanthropy. 
We thank Google for enabling experiments with Gemini through a gift.
This research was supported with Cloud TPUs from Google's TPU Research Cloud (TRC).
We thank Nathan Godey for his comments.

\bibliography{colm2026_conference}
\bibliographystyle{colm2026_conference}

\appendix
\clearpage
\appendix

\section{Additional Experimental Setup Details}
\label{app:exp}

\subsection{Data}\label{app:exp:data}
\begin{table*}[t]
\centering
\begin{tabular}{lrrr}
\toprule
Dataset & Avg. Context Tokens & \#Samples & \#Contexts \\
\midrule
\multicolumn{4}{l}{\textbf{Summarization}} \\
CNN/DM~\citep{see-etal-2017-get_cnndm}      & 649 & 198{,}732 & 196{,}601  \\
DialogSum~\citep{chen-etal-2021-dialogsum}   & 208 & 12{,}452  & 12{,}450   \\
SAMSum~\citep{Gliwa_2019_samsum}      & 145 & 14{,}730  & 14{,}254   \\
XSum~\citep{narayan-etal-2018-dont_xsum}        & 408 & 185{,}760 & 185{,}566  \\
\addlinespace
\multicolumn{4}{l}{\textbf{Reading Comprehension}} \\
BoolQ~\citep{clark-etal-2019-boolq}       & 126 & 9{,}427   & 7{,}927    \\
DROP~\citep{dua-etal-2019-drop}        & 295 & 76{,}751  & 5{,}477    \\
HotpotQA~\citep{yang-etal-2018-hotpotqa}    & 247 & 90{,}327  & 84{,}705   \\
NarrativeQA~\citep{kocisky-etal-2018-narrativeqa} & 668 & 28{,}299  & 953        \\
PubMedQA~\citep{jin-etal-2019-pubmedqa}    & 318 & 211{,}218 & 211{,}164  \\
QuAC~\citep{choi-etal-2018-quac}        & 515 & 81{,}391  & 6{,}574    \\
QuAIL~\citep{rogers2020getting_quail}       & 416 & 10{,}246  & 560        \\
RACE~\citep{lai-etal-2017-race}        & 349 & 87{,}749  & 25{,}108   \\
SQuAD~\citep{rajpurkar-etal-2016-squad}       & 162 & 86{,}821  & 18{,}877   \\
PWC~\citep{ge2024incontext}         & 477 & 241{,}563 & 16{,}382   \\
\midrule
\textbf{Total} & \textbf{410} & \textbf{1,335,466} & \textbf{786,598}  \\
\bottomrule
\end{tabular}
\caption{Training datasets with average context length (tokens), number of samples, number of distinct contexts, and task category. The overall average context length is weighted by the number of samples.}
\label{tab:training_datasets}
\end{table*}

\autoref{tab:training_datasets} provides a detailed list of our training data mixture. Evaluation dataset statistics appear in \autoref{tab:evaluation_datasets} and \autoref{tab:longbench-datasets} in the main text.
We use the summaries as contexts for NarrativeQA, instead of the full stories. For HotpotQA, we only use the two gold paragraphs as contexts, and remove the distractors. 

We increase the training data diversity by randomly sampling a prompt template that fits the task, when training samples are composed of a context $C$, question $Q$ and answer $A$. For example, for the extractive QA task, an example of a prompt template is: \textit{``\textless C\textgreater \textbackslash n Extract the answer from the text above. \textbackslash n Question: \textless Q\textgreater \textbackslash n Answer:  \textless A\textgreater''}. Similar templates are defined for other tasks as well. 
We created approximately 100 prompt templates for each task. The full list of templates for each task is available along with our code. 

\subsection{Training Hyperparameters}\label{app:exp:hps}
We ran initial hyperparameter exploration experiments using a text continuation task on a subset of the Dolma~\citep{soldaini-etal-2024-dolma} dataset. We generally found that most hyperparameters did not significantly affect perplexity on a held-out evaluation set (a different Dolma subset), except for the learning rate, which had more substantial effects. We determined the number of steps based on our computational budget and the plateauing of the loss curve. We repeated this process for each compression method and chose our final set of hyperparameters to be identical across all methods, as we found them to be near-optimal for all methods without statistically significant differences. We provide the final hyperparameters used to train all models, including the teacher and compressor models, in \autoref{table:hyperparameters}.
\begin{table}[H]
    \centering
    \begin{tabular}{cc}
        \toprule
        \textbf{Hyperparameter} & \textbf{Value}  \\
        \midrule
        LoRA $r$ & 16 \\
        LoRA $\alpha$ & 16 \\ 
        optimizer &  AdamW \\
        $\beta_1$ & 0.9 \\
        $\beta_2$ & 0.95 \\
        clip norm & 1 \\
        peak learning rate & 2e-4 \\
        final learning rate & 2e-5 \\
        lr scheduler type & cosine \\
        warmup ratio & 0.05 \\
        weight decay & 0.0 \\
        steps & 48,000 \\
        batch size & 32 \\
        max context length & 1024 \\
        max answer tokens & 256 \\
        \bottomrule
    \end{tabular}
    \caption{Hyperparameters for training all the models used in this work.} 
    \label{table:hyperparameters}
\end{table}

\subsection{Long Context Experiments}\label{app:exp:long}

\paragraph{Data} We use the training datasets listed in \autoref{tab:long_context_training_datasets} for long-context training. To preserve performance on short contexts, following \citet{chen2024longlora}, we mix these with 2,000 samples from each of the original training datasets (\autoref{tab:training_datasets}). For evaluation, we use the LongBench-E variant of LongBench~\citep{bai2024longbench}, which contains more samples with context lengths below 8K tokens; dataset statistics are provided in \autoref{tab:longbench-datasets}.

\paragraph{Training Procedure} We employ a three-stage training procedure. First, the teacher model from the 1K-context experiments is further finetuned on the long-context data mixture, adopting a progressive training strategy~\citep{qwen2025qwen25technicalreport}. Next, a compressor is trained using this teacher alongside the original 1K-context data mixture. Finally, the compressor undergoes additional training on the same long-context mixture used during teacher finetuning. We use the same hyperparameters as in \autoref{table:hyperparameters}, with two changes: (1) we reduce the number of steps to 4,800, and (2) we use a max context length of 8,192.

\paragraph{Model Choice} We run the long-context experiments with Qwen3-1.7B since it was pretrained with a 32K context length and fits within our computational budget. Gemma2-2B, while of comparable size and with better performance, was pretrained with a context length of 8K, which is insufficient given that the contexts alone in our experiments reach 8K tokens, not including prompt and answer tokens.

\begin{table*}[t]
\centering
\begin{tabular}{lrrr}
\toprule
Dataset & Avg. Context Tokens & \#Samples & \#Contexts \\
\midrule
BillSum~\citep{kornilova-eidelman-2019-billsum}         & 2{,}278 & 5{,}000 & 5{,}000   \\
HotpotQA (all contexts)~\citep{yang-etal-2018-hotpotqa}         & 1{,}338 & 5{,}000 & 5{,}000   \\
BookSum~\citep{kryscinski-etal-2022-booksum}         & 3{,}670 & 4{,}055 & 3{,}430   \\
LongAlpaca~\citep{chen2024longlora}         & 6{,}943 & 3{,}918 & 3{,}564   \\
QuALITY~\citep{pang-etal-2022-quality}         & 5{,}830 & 2{,}426 & 144   \\
QASPER~\citep{dasigi-etal-2021-qasper}         & 4{,}756 & 2{,}331 & 769   \\
QMSum~\citep{zhong-etal-2021-qmsum}         & 5{,}749 & 295 & 42   \\
\midrule
\textbf{Total} & \textbf{3,782} & \textbf{23,025} & \textbf{17,949}  \\
\bottomrule
\end{tabular}
\caption{Long context training datasets with average context length (tokens), number of samples, number of distinct contexts, and task category. The overall average context length is weighted by the number of samples.}
\label{tab:long_context_training_datasets}
\end{table*}

\subsection{Computational Resources}\label{app:exp:compute}
All experiments in this paper (except for the baseline systems) were trained and evaluated on Google Cloud preemptible TPUs, and implemented using the JAX and Flax NNX libraries. Since training was only done on preemptible TPUs, it is hard to estimate the total training time for each experiment, as most of them were interrupted several times by preemption. As rough estimates, when using a v4-64 TPU and a 2B model trained for 48,000 steps and a batch size of 32: training a teacher model took 4 hours, training a multi-ratio compressor model took 23 hours, and training a single-ratio compressor model took 10 hours.

\section{In-Domain vs. Out-of-Domain Experiments}
\label{app:in_vs_out}
\begin{figure}[t]
  \centering
  \begin{subfigure}[t]{1.\linewidth}
    \centering
    \includegraphics[width=\linewidth]{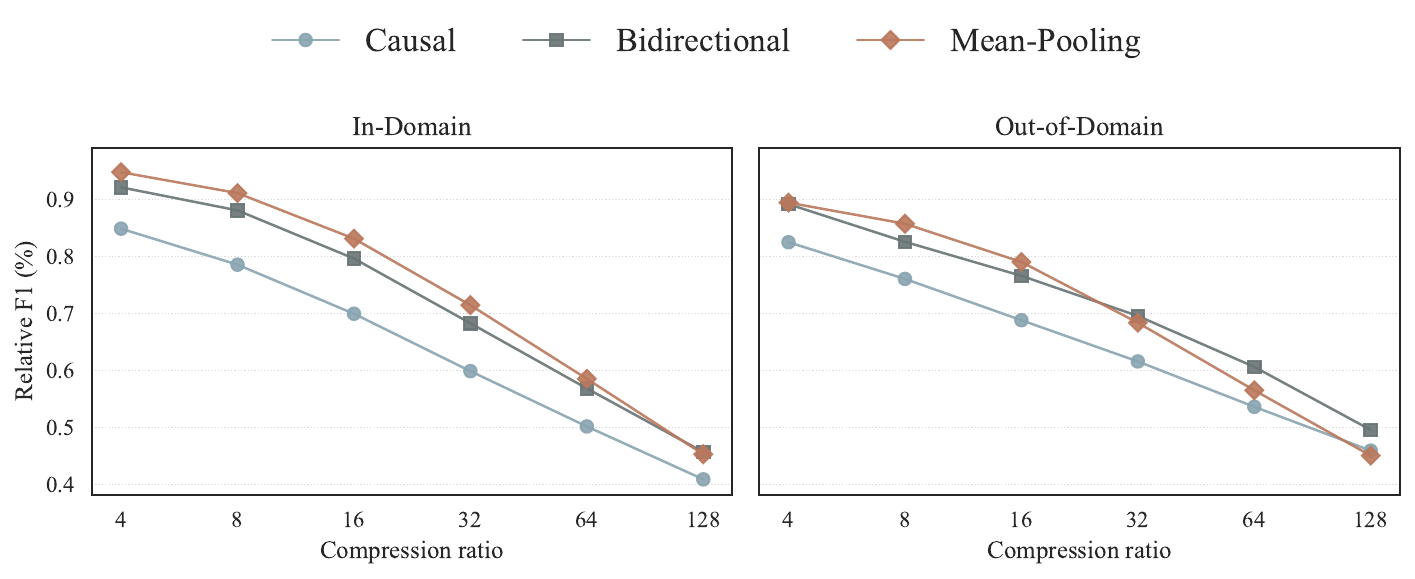}
    \caption{In-domain vs.\ out-of-domain performance across compression ratios.}
    \label{fig:in-vs-out}
  \end{subfigure}

  \vspace{0.8em} %

  \begin{subfigure}[t]{1.\linewidth}
    \centering
    \includegraphics[width=\linewidth]{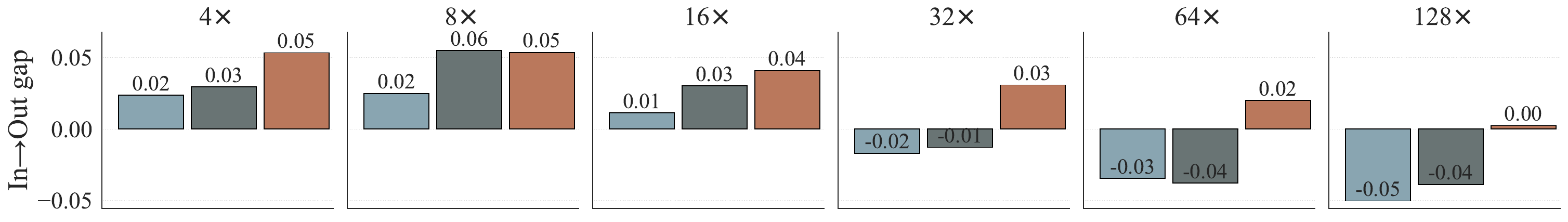}
    \caption{Performance drop (in–out gap) per method across compression ratios (in teacher-normalized Relative $F_1$ units). Higher values mean a larger domain performance gap. Negative values mean that the out-of-domain performance is better than in-domain performance.}
    \label{fig:in-out-gap}
  \end{subfigure}

  \caption{In-domain and out-of-domain comparison. (a) Line plots show performance on in-domain vs.\ out-of-domain datasets. (b) Bar plots show the in–out performance gap per method, which is the difference between in-domain and out-of-domain teacher-normalized $F_1$ scores.}
  \label{fig:in-out-combined}
\end{figure}

We construct \evalname with both in-domain QA datasets and out-of-domain QA datasets (\autoref{sec:eval:data}). The training splits of the in-domain datasets are included in the training data mixture, while the out-of-domains datasets are not. 
It is expected that  downstream performance will drop for out-of-domain datasets. 
Critical for our study, though, is the performance drop of the compressor itself. 
\autoref{fig:in-vs-out} plots the in-domain and out-of-domain performance using the teacher-normalized $F_1$ score for the Qwen3-8B model, averaged over the datasets in each category. 
We first observe that while the mean-pooling approach is superior for ratios up to $16\times$ in all settings, its performance deteriorates as the compression ratio increases. 
To better understand the performance change due to the domain gap, we plot the differences between the in-domain and out-of-domain performance in \autoref{fig:in-out-gap}. 
The performance gap is higher for low ratios, and lower for higher ratios. 
One possible explanation is that at low compression ratios the compressed representations still retain much of the original contextual signal, so the model is more sensitive to domain-specific distributional shifts; differences between in-domain and out-of-domain language patterns thus manifest as a larger performance gap. By contrast, at higher compression ratios much of the fine-grained contextual detail is already lost to compression noise, which dominates over the domain gap. In this regime, both in-domain and out-of-domain datasets suffer similarly from the limited representational capacity, resulting in a smaller relative difference. 

\FloatBarrier
\section{Compression Scaling}\label{app:scaling}

\begin{figure}[htb]
  \centering
  \includegraphics[width=1.0\linewidth]{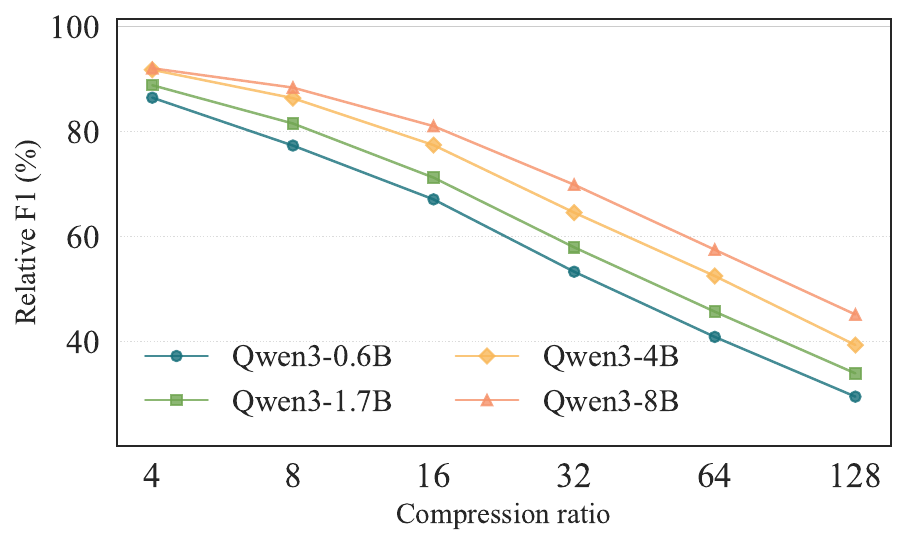}
  \caption{Compression Scaling. We show the teacher-normalized $F_1$ scores (Relative F1) across four Qwen3 model scales. The scores are averages of the scores of all datasets. We can clearly observe the benefits of scaling for LLM compressors.}
  \label{fig:scaling_law}
\end{figure}

LLM performance increases with scale~\citep{hoffmann2022training_chinchilla}, but does compression quality scale similarly? A compressor improving at the same rate as its teacher would show constant teacher-normalized scores across scales. \autoref{fig:scaling_law} shows results for four Qwen3 scales under multi-ratio training. Compressors demonstrate desirable scaling: teacher-normalized $F_1$ increases with model size, meaning the efficiency gains of compression are larger for larger models. Both baselines show this trend (\autoref{tab:primary_results_hybrid_transpose}).

\FloatBarrier
\section{Mean-Pooling Ablations}\label{app:ablations}

\begin{table}[t]
\centering
\resizebox{1.\linewidth}{!}{%
\begin{tabular}{l r r r r r r c}
\toprule
\textbf{Ablation} (\textsc{Gemma2-2B}) & 4$\times$ & 8$\times$ & 16$\times$ & 32$\times$ & 64$\times$ & 128$\times$ & $\Delta$ \\
\midrule
Default & \textbf{68.1} & \textbf{65.4} & \textbf{61.0} & \textbf{54.9} & \textbf{48.8} & \textbf{43.7} & \,($+0.0$) \\
Fixed Decoder & 64.9 & 61.9 & 57.0 & 51.5 & 45.0 & 39.8 & \,($-3.6$) \\
Fixed Encoder & 57.4 & 49.9 & 44.1 & 39.8 & 36.2 & 34.8 & \,($-13.3$) \\
No Encoder & 58.7 & 51.9 & 44.9 & 40.2 & 36.2 & 34.1 & \,($-12.6$) \\
w/o Linear Layer & 67.7 & 64.5 & 60.0 & 54.1 & 48.1 & 43.2 & \,($-0.7$) \\
Ratio Sampling & 67.1 & 64.0 & 59.3 & 53.5 & 47.5 & 42.2 & \,($-1.4$) \\
\bottomrule
\end{tabular}
}
\caption{Ablation study for mean pooling using  \textsc{Gemma2-2B} as the teacher LLM. Numbers are macro-averaged $F_1$ scores. \emph{$\Delta$}: mean change vs. Default across ratios; \textbf{bold} = best per column.}
\label{tab:abl_gemma2_mean}
\end{table}

We ablate the mean-pooling baseline using Gemma2-2B (\autoref{tab:abl_gemma2_mean}), testing: (1) Fixed Decoder (only encoder trained); (2) Fixed Encoder (only decoder trained); (3) No Encoder (context represented via decoder token embeddings only); (4) w/o Linear Layer (pooling output passed directly); (5) Ratio Sampling (one ratio sampled per instance rather than training on all ratios).
Freezing the decoder causes considerable but not catastrophic degradation, consistent with prior work~\citep{louis-etal-2025-pisco}. Freezing or removing the encoder is more detrimental ($>$12\% drop). The linear layer has minimal impact (0.7\% reduction when removed). Ratio sampling speeds up training at a small cost (1.4\% drop).

\FloatBarrier
\section{Additional Results}\label{app:results}

\begin{table*}[t]
    \small
    \centering
    \setlength{\tabcolsep}{5pt}
    \resizebox{\textwidth}{!}{%
\begin{tabular}{llcccccccc}
\toprule
\multicolumn{2}{c}{} & \multicolumn{1}{c}{\textbf{Original}} & \multicolumn2{c}{\textbf{4x}} & \multicolumn2{c}{\textbf{16x}} & \multicolumn2{c}{\textbf{128x}} & \multicolumn{1}{c}{\textbf{No Ctx}} \\
\cmidrule(lr){3-3}\cmidrule(lr){4-5}\cmidrule(lr){6-7}\cmidrule(lr){8-9}\cmidrule(l){10-10}
\multicolumn{2}{c}{} &  & \textbf{Single} & \textbf{Multi} & \textbf{Single} & \textbf{Multi} & \textbf{Single} & \textbf{Multi} &  \\
\midrule
\multicolumn{10}{l}{\textbf{Baseline Systems}} \\
& \emph{LLMLingua2} (Qwen3-8B) &  & & 30.02 &  & 16.16 &  & 16.06 &  \\
& \emph{ICAE} (Mistral-7B) &  & 24.94 &  &  &  &  &  &  \\
& \emph{PCC Lite} (GPT2-Large \& Llama3.1-8B) &  & 48.81 &  & 38.38 &  & 25.94 &  &  \\
& \emph{PCC Large} (Llama3.1-8B) &  & 49.34 &  & 36.64 &  & 27.92 &  &  \\
\midrule
\multicolumn{10}{l}{\textbf{Our Baselines}} \\
\addlinespace[2pt]
\multicolumn{2}{l}{\textbf{Qwen3-8B}} & 59.82 &  &  &  &  &  &  & 16.19 \\
 & Compression-Tokens (Causal) &  & 51.59 & 50.01 & 41.26 & 42.99 & 34.21 & 32.50 &  \\
 & Compression-Tokens (Bidirectional) &  & 53.44 & 53.99 & 44.92 & 47.15 & 33.60 & 34.27 &  \\
 & Mean-Pooling &  & 56.41 & 55.04 & 47.95 & 49.02 & 34.72 & 33.15 &  \\
\midrule
\multicolumn{2}{l}{\textbf{Qwen3-4B}} & 58.87 &  &  &  &  &  &  & 13.74 \\
 & Compression-Tokens (Causal) &  & 49.66 & 46.26 & 40.05 & 39.59 & 30.07 & 28.84 &  \\
 & Compression-Tokens (Bidirectional) &  & 51.06 & 52.37 & 42.53 & 45.14 & 28.84 & 29.51 &  \\
 & Mean-Pooling &  & 55.25 & 53.77 & 45.43 & 45.86 & 30.91 & 28.38 &  \\
\midrule
\multicolumn{2}{l}{\textbf{Qwen3-1.7B}} & 55.19 &  &  &  &  &  &  & 9.07 \\
 & Compression-Tokens (Causal) &  & 36.66 & 41.64 & 35.49 & 34.64 & 24.27 & 24.04 &  \\
 & Compression-Tokens (Bidirectional) &  & 46.45 & 46.72 & 36.62 & 38.93 & 24.31 & 24.33 &  \\
 & Mean-Pooling &  & 51.28 & 48.77 & 40.45 & 39.04 & 25.15 & 22.01 &  \\
\midrule
\multicolumn{2}{l}{\textbf{Qwen3-0.6B}} & 50.85 &  &  &  &  &  &  & 4.78 \\
 & Compression-Tokens (Causal) &  & 39.66 & 36.76 & 27.99 & 29.00 & 18.23 & 18.20 &  \\
 & Compression-Tokens (Bidirectional) &  & 40.98 & 41.92 & 30.92 & 33.64 & 18.82 & 18.55 &  \\
 & Mean-Pooling &  & 45.82 & 43.07 & 32.50 & 33.09 & 19.05 & 16.07 &  \\
\midrule
\multicolumn{2}{l}{\textbf{Gemma2-2B}} & 57.63 &  &  &  &  &  &  & 15.00 \\
 & Compression-Tokens (Causal) &  & 47.90 & 45.98 & 39.57 & 39.74 & 32.02 & 29.66 &  \\
 & Compression-Tokens (Bidirectional) &  & 49.43 & 49.40 & 40.89 & 42.70 & 31.79 & 30.43 &  \\
 & Mean-Pooling &  & 54.20 & 52.77 & 45.88 & 45.68 & 32.41 & 30.83 &  \\
\midrule
\multicolumn{2}{l}{\textbf{Llama3.2-1B}} & 51.67 &  &  &  &  &  &  & 9.47 \\
 & Compression-Tokens (Causal) &  & 41.84 & 39.03 & 33.73 & 33.38 & 24.11 & 24.64 &  \\
 & Compression-Tokens (Bidirectional) &  & 43.46 & 42.96 & 35.04 & 35.46 & 24.91 & 24.56 &  \\
 & Mean-Pooling &  & 47.97 & 45.45 & 33.15 & 36.93 & 22.89 & 22.70 &  \\
\midrule
\bottomrule
\end{tabular}
    }
    \caption{Primary results with exact match (EM) as the metric.}
    \label{tab:primary_results_EM}
\end{table*}

\begin{table*}[t]
    \small
    \centering
    \setlength{\tabcolsep}{5pt}
    \resizebox{\textwidth}{!}{%
\begin{tabular}{llcccccccc}
\toprule
\multicolumn{2}{c}{} & \multicolumn{1}{c}{\textbf{Original}} & \multicolumn2{c}{\textbf{4x}} & \multicolumn2{c}{\textbf{16x}} & \multicolumn2{c}{\textbf{128x}} & \multicolumn{1}{c}{\textbf{No Ctx}} \\
\cmidrule(lr){3-3}\cmidrule(lr){4-5}\cmidrule(lr){6-7}\cmidrule(lr){8-9}\cmidrule(l){10-10}
\multicolumn{2}{c}{} &  & \textbf{Single} & \textbf{Multi} & \textbf{Single} & \textbf{Multi} & \textbf{Single} & \textbf{Multi} &  \\
\midrule
\multicolumn{10}{l}{\textbf{Baseline Systems}} \\
& \emph{LLMLingua2} (Qwen3-8B) &  & & 33.63 &  & 18.30 &  & 17.36 &  \\
& \emph{ICAE} (Mistral-7B) &  & 49.18 &  &  &  &  &  &  \\
& \emph{PISCO} (Llama3.1-8B) &  &  &  & 53.62 &  &  &  &  \\
& \emph{PCC Lite} (GPT2-Large \& Llama3.1-8B) &  & 54.05 &  & 43.67 &  & 30.03 &  &  \\
& \emph{PCC Large} (Llama3.1-8B) &  & 55.17 &  & 41.79 &  & 30.10 &  &  \\
\midrule
\multicolumn{10}{l}{\textbf{Our Baselines}} \\
\addlinespace[2pt]
\multicolumn{2}{l}{\textbf{Qwen3-8B}} & 68.84 &  &  &  &  &  &  & 17.98 \\
 & Compression-Tokens (Causal) &  & 59.79 & 58.58 & 47.07 & 49.99 & 39.09 & 37.05 &  \\
 & Compression-Tokens (Bidirectional) &  & 62.50 & 63.58 & 52.22 & 55.26 & 38.72 & 39.36 &  \\
 & Mean-Pooling &  & 65.91 & 65.06 & 55.68 & 56.77 & 39.95 & 37.80 &  \\
\midrule
\multicolumn{2}{l}{\textbf{Qwen3-4B}} & 67.69 &  &  &  &  &  &  & 15.00 \\
 & Compression-Tokens (Causal) &  & 57.12 & 54.17 & 46.06 & 45.47 & 34.83 & 33.48 &  \\
 & Compression-Tokens (Bidirectional) &  & 59.35 & 60.99 & 48.92 & 52.23 & 33.20 & 34.33 &  \\
 & Mean-Pooling &  & 64.05 & 62.94 & 52.77 & 52.82 & 35.49 & 32.37 &  \\
\midrule
\multicolumn{2}{l}{\textbf{Qwen3-1.7B}} & 64.21 &  &  &  &  &  &  & 9.85 \\
 & Compression-Tokens (Causal) &  & 43.01 & 49.31 & 41.31 & 40.45 & 28.45 & 27.90 &  \\
 & Compression-Tokens (Bidirectional) &  & 54.31 & 54.70 & 42.66 & 45.13 & 28.33 & 28.04 &  \\
 & Mean-Pooling &  & 60.03 & 57.28 & 46.63 & 45.07 & 28.96 & 25.82 &  \\
\midrule
\multicolumn{2}{l}{\textbf{Qwen3-0.6B}} & 59.67 &  &  &  &  &  &  & 5.62 \\
 & Compression-Tokens (Causal) &  & 46.36 & 43.62 & 33.07 & 34.30 & 21.13 & 21.71 &  \\
 & Compression-Tokens (Bidirectional) &  & 48.04 & 48.84 & 36.11 & 39.50 & 22.13 & 22.00 &  \\
 & Mean-Pooling &  & 54.36 & 51.16 & 38.30 & 38.68 & 22.44 & 18.87 &  \\
\midrule
\multicolumn{2}{l}{\textbf{Gemma2-2B}} & 66.14 &  &  &  &  &  &  & 16.80 \\
 & Compression-Tokens (Causal) &  & 55.50 & 52.94 & 46.13 & 45.59 & 36.14 & 33.94 &  \\
 & Compression-Tokens (Bidirectional) &  & 56.94 & 57.54 & 46.90 & 49.52 & 36.14 & 34.75 &  \\
 & Mean-Pooling &  & 62.51 & 61.28 & 52.55 & 51.90 & 36.61 & 34.93 &  \\
\midrule
\multicolumn{2}{l}{\textbf{Llama3.2-1B}} & 60.30 &  &  &  &  &  &  & 11.09 \\
 & Compression-Tokens (Causal) &  & 48.85 & 45.41 & 39.16 & 38.51 & 27.54 & 28.01 &  \\
 & Compression-Tokens (Bidirectional) &  & 50.89 & 50.10 & 40.44 & 41.91 & 28.34 & 28.44 &  \\
 & Mean-Pooling &  & 56.62 & 53.50 & 38.84 & 42.76 & 25.99 & 26.19 &  \\
\midrule
\bottomrule
\end{tabular}
    }
    \caption{Primary results with accuracy as the metric.}
    \label{tab:primary_results_accuracy}
\end{table*}

We provide additional results from the same experiments conducted in the main body of the paper. \aautoref{app:results:metrics} provides the primary results of the paper with the EM and substring accuracy metrics. \aautoref{app:results:datasets} shows the $F_1$ performance on each individual dataset from the evaluation suite. 

\subsection{Primary Results \textemdash Additional Metrics}\label{app:results:metrics}

We provide our primary results with the EM and accuracy metrics in \autoref{tab:primary_results_EM} and \autoref{tab:primary_results_accuracy} respectively, akin to those presented in \autoref{tab:primary_results_hybrid_transpose}. 

\subsection{Results Per Dataset}\label{app:results:datasets}

Our evaluation suite comprises six datasets. Here we present individual dataset performance using the $F_1$ metric.
\subsubsection{In Domain Datasets Results}\label{app:results:datasets:in}
We provide results for SQuAD, HotpotQA and NarrativeQA in \autoref{tab:primary_results_squad}, \autoref{tab:primary_results_hotpotqa} and \autoref{tab:primary_results_narrativeqa}, respectively. 
\begin{table*}[t]
    \small
    \centering
    \setlength{\tabcolsep}{5pt}
    \resizebox{\textwidth}{!}{%
\begin{tabular}{llcccccccc}
\toprule
\multicolumn{2}{c}{} & \multicolumn{1}{c}{\textbf{Original}} & \multicolumn2{c}{\textbf{4x}} & \multicolumn2{c}{\textbf{16x}} & \multicolumn2{c}{\textbf{128x}} & \multicolumn{1}{c}{\textbf{No Ctx}} \\
\cmidrule(lr){3-3}\cmidrule(lr){4-5}\cmidrule(lr){6-7}\cmidrule(lr){8-9}\cmidrule(l){10-10}
\multicolumn{2}{c}{} &  & \textbf{Single} & \textbf{Multi} & \textbf{Single} & \textbf{Multi} & \textbf{Single} & \textbf{Multi} &  \\
\midrule
\multicolumn{10}{l}{\textbf{Baseline Systems}} \\
& \emph{LLMLingua2} (Qwen3-8B) &  & & 48.38 &  & 21.34 &  & 19.68 &  \\
& \emph{ICAE} (Mistral-7B) &  & 45.6 &  &  &  &  &  &  \\
& \emph{PCC Lite} (GPT2-Large \& Llama3.1-8B) &  & 78.38 &  & 67.63 &  & 40.22 &  &  \\
& \emph{PCC Large} (Llama3.1-8B) &  & 79.56 &  & 62.93 &  & 41.23 &  &  \\
\midrule
\multicolumn{10}{l}{\textbf{Our Baselines}} \\
\addlinespace[2pt]
\multicolumn{2}{l}{\textbf{Qwen3-8B}} & 86.48 &  &  &  &  &  &  & 20.31 \\
 & Compression-Tokens (Causal) &  & 77.11 & 74.89 & 57.05 & 62.12 & 44.56 & 42.35 &  \\
 & Compression-Tokens (Bidirectional) &  & 80.00 & 81.23 & 64.80 & 69.27 & 44.30 & 43.86 &  \\
 & Mean-Pooling &  & 83.76 & 82.75 & 71.37 & 71.19 & 44.65 & 43.19 &  \\
\midrule
\multicolumn{2}{l}{\textbf{Qwen3-4B}} & 85.75 &  &  &  &  &  &  & 17.72 \\
 & Compression-Tokens (Causal) &  & 74.23 & 71.31 & 57.49 & 56.88 & 38.95 & 37.10 &  \\
 & Compression-Tokens (Bidirectional) &  & 77.24 & 79.28 & 60.71 & 65.49 & 38.19 & 38.41 &  \\
 & Mean-Pooling &  & 83.19 & 81.54 & 68.20 & 67.47 & 39.96 & 37.54 &  \\
\midrule
\multicolumn{2}{l}{\textbf{Qwen3-1.7B}} & 83.65 &  &  &  &  &  &  & 12.66 \\
 & Compression-Tokens (Causal) &  & 54.25 & 64.50 & 49.09 & 49.98 & 31.78 & 30.10 &  \\
 & Compression-Tokens (Bidirectional) &  & 72.92 & 73.39 & 53.07 & 57.36 & 31.58 & 31.17 &  \\
 & Mean-Pooling &  & 79.56 & 77.17 & 59.01 & 58.30 & 32.36 & 29.90 &  \\
\midrule
\multicolumn{2}{l}{\textbf{Qwen3-0.6B}} & 81.55 &  &  &  &  &  &  & 7.91 \\
 & Compression-Tokens (Causal) &  & 61.67 & 57.52 & 40.29 & 41.60 & 22.06 & 22.45 &  \\
 & Compression-Tokens (Bidirectional) &  & 64.21 & 67.05 & 42.81 & 46.68 & 22.09 & 22.36 &  \\
 & Mean-Pooling &  & 74.00 & 70.60 & 48.62 & 48.14 & 22.40 & 20.45 &  \\
\midrule
\multicolumn{2}{l}{\textbf{Gemma2-2B}} & 84.58 &  &  &  &  &  &  & 16.41 \\
 & Compression-Tokens (Causal) &  & 70.61 & 69.06 & 55.75 & 56.37 & 37.89 & 35.66 &  \\
 & Compression-Tokens (Bidirectional) &  & 74.16 & 75.41 & 57.77 & 61.75 & 37.72 & 37.00 &  \\
 & Mean-Pooling &  & 81.67 & 80.01 & 66.38 & 65.64 & 38.96 & 36.71 &  \\
\midrule
\multicolumn{2}{l}{\textbf{Llama3.2-1B}} & 81.16 &  &  &  &  &  &  & 11.27 \\
 & Compression-Tokens (Causal) &  & 62.65 & 59.34 & 46.41 & 45.49 & 28.58 & 28.44 &  \\
 & Compression-Tokens (Bidirectional) &  & 64.48 & 65.61 & 48.99 & 51.39 & 29.09 & 28.82 &  \\
 & Mean-Pooling &  & 74.91 & 71.40 & 47.36 & 53.66 & 28.02 & 27.91 &  \\
\midrule
\bottomrule
\end{tabular}
    }
    \caption{SQuAD $F_1$.}
    \label{tab:primary_results_squad}
\end{table*}

\begin{table*}[t]
    \small
    \centering
    \setlength{\tabcolsep}{5pt}
    \resizebox{\textwidth}{!}{%
\begin{tabular}{llcccccccc}
\toprule
\multicolumn{2}{c}{} & \multicolumn{1}{c}{\textbf{Original}} & \multicolumn2{c}{\textbf{4x}} & \multicolumn2{c}{\textbf{16x}} & \multicolumn2{c}{\textbf{128x}} & \multicolumn{1}{c}{\textbf{No Ctx}} \\
\cmidrule(lr){3-3}\cmidrule(lr){4-5}\cmidrule(lr){6-7}\cmidrule(lr){8-9}\cmidrule(l){10-10}
\multicolumn{2}{c}{} &  & \textbf{Single} & \textbf{Multi} & \textbf{Single} & \textbf{Multi} & \textbf{Single} & \textbf{Multi} &  \\
\midrule
\multicolumn{10}{l}{\textbf{Baseline Systems}} \\
& \emph{LLMLingua2} (Qwen3-8B) &  & & 53.54 &  & 29.32 &  & 26.27 &  \\
& \emph{ICAE} (Mistral-7B) &  & 50.01 &  &  &  &  &  &  \\
& \emph{PCC Lite} (GPT2-Large \& Llama3.1-8B) &  & 68.55 &  & 59.38 &  & 43.93 &  &  \\
& \emph{PCC Large} (Llama3.1-8B) &  & 70.08 &  & 59.05 &  & 46.46 &  &  \\
\midrule
\multicolumn{10}{l}{\textbf{Our Baselines}} \\
\addlinespace[2pt]
\multicolumn{2}{l}{\textbf{Qwen3-8B}} & 84.67 &  &  &  &  &  &  & 26.65 \\
 & Compression-Tokens (Causal) &  & 78.85 & 78.32 & 68.00 & 72.65 & 64.14 & 59.74 &  \\
 & Compression-Tokens (Bidirectional) &  & 80.24 & 81.45 & 73.30 & 76.77 & 63.26 & 62.44 &  \\
 & Mean-Pooling &  & 83.30 & 82.08 & 77.66 & 78.41 & 63.88 & 63.77 &  \\
\midrule
\multicolumn{2}{l}{\textbf{Qwen3-4B}} & 84.12 &  &  &  &  &  &  & 23.16 \\
 & Compression-Tokens (Causal) &  & 76.48 & 75.56 & 68.85 & 69.80 & 59.08 & 55.78 &  \\
 & Compression-Tokens (Bidirectional) &  & 78.13 & 79.41 & 71.11 & 74.60 & 58.38 & 56.87 &  \\
 & Mean-Pooling &  & 82.20 & 80.77 & 75.45 & 76.02 & 59.29 & 58.74 &  \\
\midrule
\multicolumn{2}{l}{\textbf{Qwen3-1.7B}} & 80.95 &  &  &  &  &  &  & 18.75 \\
 & Compression-Tokens (Causal) &  & 66.69 & 70.98 & 63.82 & 63.84 & 51.48 & 48.78 &  \\
 & Compression-Tokens (Bidirectional) &  & 73.73 & 74.93 & 66.06 & 68.50 & 52.08 & 50.77 &  \\
 & Mean-Pooling &  & 78.64 & 76.11 & 68.76 & 68.78 & 50.86 & 49.44 &  \\
\midrule
\multicolumn{2}{l}{\textbf{Qwen3-0.6B}} & 77.35 &  &  &  &  &  &  & 14.74 \\
 & Compression-Tokens (Causal) &  & 66.62 & 65.76 & 55.88 & 57.79 & 43.16 & 39.58 &  \\
 & Compression-Tokens (Bidirectional) &  & 67.24 & 69.28 & 58.44 & 61.79 & 43.80 & 41.66 &  \\
 & Mean-Pooling &  & 73.00 & 69.58 & 61.13 & 61.08 & 42.76 & 40.45 &  \\
\midrule
\multicolumn{2}{l}{\textbf{Gemma2-2B}} & 82.55 &  &  &  &  &  &  & 25.18 \\
 & Compression-Tokens (Causal) &  & 75.62 & 75.54 & 69.18 & 70.42 & 61.91 & 59.03 &  \\
 & Compression-Tokens (Bidirectional) &  & 76.55 & 77.60 & 69.64 & 73.51 & 61.67 & 60.46 &  \\
 & Mean-Pooling &  & 80.93 & 79.85 & 75.10 & 74.90 & 62.36 & 61.64 &  \\
\midrule
\multicolumn{2}{l}{\textbf{Llama3.2-1B}} & 77.96 &  &  &  &  &  &  & 19.34 \\
 & Compression-Tokens (Causal) &  & 69.68 & 68.22 & 63.36 & 62.96 & 52.27 & 49.81 &  \\
 & Compression-Tokens (Bidirectional) &  & 70.74 & 71.01 & 64.47 & 66.19 & 53.51 & 51.79 &  \\
 & Mean-Pooling &  & 74.38 & 72.69 & 62.75 & 66.59 & 51.05 & 50.86 &  \\
\midrule
\bottomrule
\end{tabular}
    }
    \caption{HotpotQA $F_1$.}
    \label{tab:primary_results_hotpotqa}
\end{table*}

\begin{table*}[t]
    \small
    \centering
    \setlength{\tabcolsep}{5pt}
    \resizebox{\textwidth}{!}{%
\begin{tabular}{llcccccccc}
\toprule
\multicolumn{2}{c}{} & \multicolumn{1}{c}{\textbf{Original}} & \multicolumn2{c}{\textbf{4x}} & \multicolumn2{c}{\textbf{16x}} & \multicolumn2{c}{\textbf{128x}} & \multicolumn{1}{c}{\textbf{No Ctx}} \\
\cmidrule(lr){3-3}\cmidrule(lr){4-5}\cmidrule(lr){6-7}\cmidrule(lr){8-9}\cmidrule(l){10-10}
\multicolumn{2}{c}{} &  & \textbf{Single} & \textbf{Multi} & \textbf{Single} & \textbf{Multi} & \textbf{Single} & \textbf{Multi} &  \\
\midrule
\multicolumn{10}{l}{\textbf{Baseline Systems}} \\
& \emph{LLMLingua2} (Qwen3-8B) &  & & 27.33 &  & 14.35 &  & 10.69 &  \\
& \emph{ICAE} (Mistral-7B) &  & 32.65 &  &  &  &  &  &  \\
& \emph{PCC Lite} (GPT2-Large \& Llama3.1-8B) &  & 50.29 &  & 34.16 &  & 16.05 &  &  \\
& \emph{PCC Large} (Llama3.1-8B) &  & 50.72 &  & 32.56 &  & 16.18 &  &  \\
\midrule
\multicolumn{10}{l}{\textbf{Our Baselines}} \\
\addlinespace[2pt]
\multicolumn{2}{l}{\textbf{Qwen3-8B}} & 68.00 &  &  &  &  &  &  & 10.93 \\
 & Compression-Tokens (Causal) &  & 59.62 & 58.28 & 46.58 & 49.32 & 33.41 & 29.37 &  \\
 & Compression-Tokens (Bidirectional) &  & 61.44 & 62.13 & 51.48 & 55.68 & 34.17 & 33.61 &  \\
 & Mean-Pooling &  & 65.89 & 64.74 & 58.17 & 58.38 & 34.81 & 32.21 &  \\
\midrule
\multicolumn{2}{l}{\textbf{Qwen3-4B}} & 67.12 &  &  &  &  &  &  & 10.40 \\
 & Compression-Tokens (Causal) &  & 57.18 & 55.08 & 46.69 & 43.77 & 30.39 & 27.84 &  \\
 & Compression-Tokens (Bidirectional) &  & 58.37 & 60.74 & 48.84 & 52.41 & 29.77 & 30.22 &  \\
 & Mean-Pooling &  & 65.22 & 63.38 & 56.14 & 55.42 & 32.66 & 28.99 &  \\
\midrule
\multicolumn{2}{l}{\textbf{Qwen3-1.7B}} & 64.42 &  &  &  &  &  &  & 7.57 \\
 & Compression-Tokens (Causal) &  & 41.97 & 49.85 & 40.15 & 39.49 & 25.42 & 23.64 &  \\
 & Compression-Tokens (Bidirectional) &  & 55.68 & 56.49 & 44.87 & 47.28 & 25.16 & 26.15 &  \\
 & Mean-Pooling &  & 60.34 & 58.56 & 48.95 & 48.78 & 26.91 & 23.60 &  \\
\midrule
\multicolumn{2}{l}{\textbf{Qwen3-0.6B}} & 61.13 &  &  &  &  &  &  & 7.76 \\
 & Compression-Tokens (Causal) &  & 48.09 & 46.30 & 34.68 & 36.05 & 20.10 & 20.69 &  \\
 & Compression-Tokens (Bidirectional) &  & 48.04 & 50.67 & 38.85 & 40.35 & 21.29 & 21.07 &  \\
 & Mean-Pooling &  & 56.48 & 53.12 & 42.47 & 42.21 & 21.29 & 19.06 &  \\
\midrule
\multicolumn{2}{l}{\textbf{Gemma2-2B}} & 66.47 &  &  &  &  &  &  & 10.34 \\
 & Compression-Tokens (Causal) &  & 56.17 & 55.78 & 47.16 & 46.68 & 31.17 & 29.72 &  \\
 & Compression-Tokens (Bidirectional) &  & 59.20 & 59.51 & 48.59 & 51.43 & 31.50 & 31.91 &  \\
 & Mean-Pooling &  & 64.29 & 63.44 & 56.94 & 55.71 & 33.42 & 31.79 &  \\
\midrule
\multicolumn{2}{l}{\textbf{Llama3.2-1B}} & 61.67 &  &  &  &  &  &  & 9.03 \\
 & Compression-Tokens (Causal) &  & 48.57 & 45.60 & 38.44 & 37.12 & 23.03 & 23.14 &  \\
 & Compression-Tokens (Bidirectional) &  & 52.24 & 49.98 & 40.65 & 42.08 & 23.59 & 23.72 &  \\
 & Mean-Pooling &  & 57.97 & 55.15 & 38.58 & 46.23 & 19.53 & 23.17 &  \\
\midrule
\bottomrule
\end{tabular}
    }
    \caption{NarrativeQA $F_1$.}
    \label{tab:primary_results_narrativeqa}
\end{table*}

\subsubsection{Out-of-Domain Datasets Results}\label{app:results:datasets:out}
We provide results for TriviaQA, AdversarialQA and ParaphraseRC in \autoref{tab:primary_results_triviaqa}, \autoref{tab:primary_results_adversarial_qa} and \autoref{tab:primary_results_paraphrase_rc}, respectively. 
\begin{table*}[t]
    \small
    \centering
    \setlength{\tabcolsep}{5pt}
    \resizebox{\textwidth}{!}{%
\begin{tabular}{llcccccccc}
\toprule
\multicolumn{2}{c}{} & \multicolumn{1}{c}{\textbf{Original}} & \multicolumn2{c}{\textbf{4x}} & \multicolumn2{c}{\textbf{16x}} & \multicolumn2{c}{\textbf{128x}} & \multicolumn{1}{c}{\textbf{No Ctx}} \\
\cmidrule(lr){3-3}\cmidrule(lr){4-5}\cmidrule(lr){6-7}\cmidrule(lr){8-9}\cmidrule(l){10-10}
\multicolumn{2}{c}{} &  & \textbf{Single} & \textbf{Multi} & \textbf{Single} & \textbf{Multi} & \textbf{Single} & \textbf{Multi} &  \\
\midrule
\multicolumn{10}{l}{\textbf{Baseline Systems}} \\
& \emph{LLMLingua2} (Qwen3-8B) &  & & 65.65 &  & 46.46 &  & 52.55 &  \\
& \emph{ICAE} (Mistral-7B) &  & 70.63 &  &  &  &  &  &  \\
& \emph{PCC Lite} (GPT2-Large \& Llama3.1-8B) &  & 86.43 &  & 78.03 &  & 72.50 &  &  \\
& \emph{PCC Large} (Llama3.1-8B) &  & 86.64 &  & 77.13 &  & 74.08 &  &  \\
\midrule
\multicolumn{10}{l}{\textbf{Our Baselines}} \\
\addlinespace[2pt]
\multicolumn{2}{l}{\textbf{Qwen3-8B}} & 89.65 &  &  &  &  &  &  & 53.79 \\
 & Compression-Tokens (Causal) &  & 89.67 & 89.15 & 88.92 & 85.50 & 79.36 & 77.55 &  \\
 & Compression-Tokens (Bidirectional) &  & 90.44 & 89.41 & 87.07 & 87.95 & 75.90 & 78.17 &  \\
 & Mean-Pooling &  & 87.94 & 86.52 & 84.38 & 86.32 & 79.74 & 75.89 &  \\
\midrule
\multicolumn{2}{l}{\textbf{Qwen3-4B}} & 90.46 &  &  &  &  &  &  & 43.49 \\
 & Compression-Tokens (Causal) &  & 88.49 & 83.28 & 82.95 & 80.42 & 72.27 & 70.84 &  \\
 & Compression-Tokens (Bidirectional) &  & 91.59 & 90.83 & 86.10 & 87.53 & 67.43 & 74.52 &  \\
 & Mean-Pooling &  & 85.50 & 88.72 & 83.68 & 85.32 & 71.04 & 67.50 &  \\
\midrule
\multicolumn{2}{l}{\textbf{Qwen3-1.7B}} & 89.20 &  &  &  &  &  &  & 25.08 \\
 & Compression-Tokens (Causal) &  & 74.89 & 83.83 & 80.55 & 73.91 & 61.72 & 64.02 &  \\
 & Compression-Tokens (Bidirectional) &  & 85.62 & 86.41 & 76.15 & 80.34 & 61.46 & 61.67 &  \\
 & Mean-Pooling &  & 85.83 & 82.75 & 80.61 & 75.94 & 63.03 & 53.77 &  \\
\midrule
\multicolumn{2}{l}{\textbf{Qwen3-0.6B}} & 81.55 &  &  &  &  &  &  & 9.87 \\
 & Compression-Tokens (Causal) &  & 78.27 & 73.57 & 62.14 & 64.79 & 48.97 & 49.69 &  \\
 & Compression-Tokens (Bidirectional) &  & 81.58 & 78.38 & 69.22 & 74.16 & 50.24 & 54.05 &  \\
 & Mean-Pooling &  & 81.45 & 77.88 & 69.08 & 70.41 & 52.45 & 43.08 &  \\
\midrule
\multicolumn{2}{l}{\textbf{Gemma2-2B}} & 90.69 &  &  &  &  &  &  & 54.06 \\
 & Compression-Tokens (Causal) &  & 89.75 & 85.06 & 82.73 & 79.19 & 77.64 & 73.71 &  \\
 & Compression-Tokens (Bidirectional) &  & 88.52 & 87.14 & 85.32 & 84.63 & 78.15 & 73.59 &  \\
 & Mean-Pooling &  & 89.09 & 86.99 & 84.95 & 85.27 & 75.30 & 74.29 &  \\
\midrule
\multicolumn{2}{l}{\textbf{Llama3.2-1B}} & 82.13 &  &  &  &  &  &  & 31.14 \\
 & Compression-Tokens (Causal) &  & 84.40 & 79.82 & 75.47 & 74.28 & 65.03 & 67.58 &  \\
 & Compression-Tokens (Bidirectional) &  & 83.72 & 84.18 & 78.94 & 73.03 & 64.83 & 68.57 &  \\
 & Mean-Pooling &  & 84.87 & 83.62 & 73.95 & 76.02 & 62.02 & 60.28 &  \\
\midrule
\bottomrule
\end{tabular}
    }
    \caption{TriviaQA Verified $F_1$.}
    \label{tab:primary_results_triviaqa}
\end{table*}

\begin{table*}[t]
    \small
    \centering
    \setlength{\tabcolsep}{5pt}
    \resizebox{\textwidth}{!}{%
\begin{tabular}{llcccccccc}
\toprule
\multicolumn{2}{c}{} & \multicolumn{1}{c}{\textbf{Original}} & \multicolumn2{c}{\textbf{4x}} & \multicolumn2{c}{\textbf{16x}} & \multicolumn2{c}{\textbf{128x}} & \multicolumn{1}{c}{\textbf{No Ctx}} \\
\cmidrule(lr){3-3}\cmidrule(lr){4-5}\cmidrule(lr){6-7}\cmidrule(lr){8-9}\cmidrule(l){10-10}
\multicolumn{2}{c}{} &  & \textbf{Single} & \textbf{Multi} & \textbf{Single} & \textbf{Multi} & \textbf{Single} & \textbf{Multi} &  \\
\midrule
\multicolumn{10}{l}{\textbf{Baseline Systems}} \\
& \emph{LLMLingua2} (Qwen3-8B) &  & & 32.79 &  & 19.56 &  & 18.76 &  \\
& \emph{ICAE} (Mistral-7B) &  & 27.34 &  &  &  &  &  &  \\
& \emph{PCC Lite} (GPT2-Large \& Llama3.1-8B) &  & 42.51 &  & 35.36 &  & 26.44 &  &  \\
& \emph{PCC Large} (Llama3.1-8B) &  & 44.09 &  & 33.39 &  & 27.52 &  &  \\
\midrule
\multicolumn{10}{l}{\textbf{Our Baselines}} \\
\addlinespace[2pt]
\multicolumn{2}{l}{\textbf{Qwen3-8B}} & 60.44 &  &  &  &  &  &  & 19.15 \\
 & Compression-Tokens (Causal) &  & 47.26 & 45.97 & 36.35 & 39.16 & 32.42 & 31.15 &  \\
 & Compression-Tokens (Bidirectional) &  & 51.46 & 51.51 & 40.05 & 42.51 & 32.54 & 32.69 &  \\
 & Mean-Pooling &  & 53.71 & 53.32 & 42.55 & 45.07 & 32.52 & 31.36 &  \\
\midrule
\multicolumn{2}{l}{\textbf{Qwen3-4B}} & 57.04 &  &  &  &  &  &  & 17.38 \\
 & Compression-Tokens (Causal) &  & 45.07 & 43.44 & 34.61 & 34.34 & 28.95 & 26.25 &  \\
 & Compression-Tokens (Bidirectional) &  & 45.35 & 47.93 & 36.32 & 38.23 & 28.17 & 27.88 &  \\
 & Mean-Pooling &  & 51.47 & 48.49 & 40.36 & 39.09 & 28.84 & 27.31 &  \\
\midrule
\multicolumn{2}{l}{\textbf{Qwen3-1.7B}} & 46.62 &  &  &  &  &  &  & 14.86 \\
 & Compression-Tokens (Causal) &  & 30.09 & 34.01 & 29.29 & 29.47 & 22.27 & 22.82 &  \\
 & Compression-Tokens (Bidirectional) &  & 37.92 & 37.00 & 29.72 & 31.44 & 22.43 & 21.05 &  \\
 & Mean-Pooling &  & 42.14 & 39.78 & 32.33 & 32.46 & 22.01 & 22.18 &  \\
\midrule
\multicolumn{2}{l}{\textbf{Qwen3-0.6B}} & 39.04 &  &  &  &  &  &  & 10.97 \\
 & Compression-Tokens (Causal) &  & 29.69 & 28.58 & 23.55 & 23.95 & 18.80 & 19.46 &  \\
 & Compression-Tokens (Bidirectional) &  & 29.16 & 33.06 & 25.47 & 26.99 & 19.60 & 18.80 &  \\
 & Mean-Pooling &  & 33.84 & 32.70 & 26.11 & 26.21 & 19.52 & 18.08 &  \\
\midrule
\multicolumn{2}{l}{\textbf{Gemma2-2B}} & 51.45 &  &  &  &  &  &  & 16.76 \\
 & Compression-Tokens (Causal) &  & 39.83 & 40.80 & 33.93 & 35.23 & 28.83 & 29.06 &  \\
 & Compression-Tokens (Bidirectional) &  & 40.76 & 41.96 & 34.73 & 35.52 & 29.47 & 27.27 &  \\
 & Mean-Pooling &  & 45.61 & 44.88 & 37.14 & 36.88 & 28.70 & 28.94 &  \\
\midrule
\multicolumn{2}{l}{\textbf{Llama3.2-1B}} & 39.75 &  &  &  &  &  &  & 14.30 \\
 & Compression-Tokens (Causal) &  & 30.58 & 29.42 & 26.27 & 26.66 & 21.04 & 21.63 &  \\
 & Compression-Tokens (Bidirectional) &  & 31.71 & 30.30 & 25.77 & 28.88 & 23.74 & 20.71 &  \\
 & Mean-Pooling &  & 34.84 & 33.60 & 27.18 & 27.29 & 21.21 & 20.46 &  \\
\midrule
\bottomrule
\end{tabular}
    }
    \caption{AdversarialQA $F_1$.}
    \label{tab:primary_results_adversarial_qa}
\end{table*}

\begin{table*}[t]
    \small
    \centering
    \setlength{\tabcolsep}{5pt}
    \resizebox{\textwidth}{!}{%
\begin{tabular}{llcccccccc}
\toprule
\multicolumn{2}{c}{} & \multicolumn{1}{c}{\textbf{Original}} & \multicolumn2{c}{\textbf{4x}} & \multicolumn2{c}{\textbf{16x}} & \multicolumn2{c}{\textbf{128x}} & \multicolumn{1}{c}{\textbf{No Ctx}} \\
\cmidrule(lr){3-3}\cmidrule(lr){4-5}\cmidrule(lr){6-7}\cmidrule(lr){8-9}\cmidrule(l){10-10}
\multicolumn{2}{c}{} &  & \textbf{Single} & \textbf{Multi} & \textbf{Single} & \textbf{Multi} & \textbf{Single} & \textbf{Multi} &  \\
\midrule
\multicolumn{10}{l}{\textbf{Baseline Systems}} \\
& \emph{LLMLingua2} (Qwen3-8B) &  & & 27.42 &  & 15.32 &  & 7.56 &  \\
& \emph{ICAE} (Mistral-7B) &  & 28.42 &  &  &  &  &  &  \\
& \emph{PCC Lite} (GPT2-Large \& Llama3.1-8B) &  & 46.31 &  & 33.25 &  & 18.14 &  &  \\
& \emph{PCC Large} (Llama3.1-8B) &  & 46.77 &  & 31.17 &  & 17.95 &  &  \\
\midrule
\multicolumn{10}{l}{\textbf{Our Baselines}} \\
\addlinespace[2pt]
\multicolumn{2}{l}{\textbf{Qwen3-8B}} & 56.77 &  &  &  &  &  &  & 7.49 \\
 & Compression-Tokens (Causal) &  & 49.69 & 48.77 & 40.37 & 41.71 & 30.94 & 28.40 &  \\
 & Compression-Tokens (Bidirectional) &  & 51.65 & 51.68 & 44.91 & 45.85 & 31.40 & 31.07 &  \\
 & Mean-Pooling &  & 55.36 & 53.87 & 48.95 & 48.63 & 31.79 & 29.12 &  \\
\midrule
\multicolumn{2}{l}{\textbf{Qwen3-4B}} & 56.14 &  &  &  &  &  &  & 6.59 \\
 & Compression-Tokens (Causal) &  & 47.85 & 46.50 & 40.73 & 40.49 & 28.86 & 27.16 &  \\
 & Compression-Tokens (Bidirectional) &  & 49.63 & 50.74 & 42.98 & 44.60 & 27.72 & 28.07 &  \\
 & Mean-Pooling &  & 54.74 & 53.24 & 46.93 & 47.01 & 29.95 & 26.25 &  \\
\midrule
\multicolumn{2}{l}{\textbf{Qwen3-1.7B}} & 54.75 &  &  &  &  &  &  & 5.09 \\
 & Compression-Tokens (Causal) &  & 37.53 & 43.23 & 36.08 & 35.38 & 24.47 & 22.71 &  \\
 & Compression-Tokens (Bidirectional) &  & 46.33 & 47.36 & 39.28 & 39.72 & 24.81 & 23.81 &  \\
 & Mean-Pooling &  & 52.04 & 50.67 & 42.90 & 42.55 & 25.16 & 22.02 &  \\
\midrule
\multicolumn{2}{l}{\textbf{Qwen3-0.6B}} & 51.54 &  &  &  &  &  &  & 4.82 \\
 & Compression-Tokens (Causal) &  & 42.04 & 39.38 & 32.91 & 31.36 & 20.07 & 19.73 &  \\
 & Compression-Tokens (Bidirectional) &  & 43.34 & 43.74 & 34.15 & 35.77 & 21.12 & 19.16 &  \\
 & Mean-Pooling &  & 48.26 & 46.29 & 38.13 & 37.81 & 21.24 & 17.03 &  \\
\midrule
\multicolumn{2}{l}{\textbf{Gemma2-2B}} & 56.00 &  &  &  &  &  &  & 7.10 \\
 & Compression-Tokens (Causal) &  & 48.10 & 46.86 & 41.66 & 40.33 & 29.30 & 27.74 &  \\
 & Compression-Tokens (Bidirectional) &  & 49.36 & 49.83 & 42.30 & 43.72 & 29.85 & 28.81 &  \\
 & Mean-Pooling &  & 54.39 & 53.39 & 47.82 & 47.83 & 31.16 & 28.91 &  \\
\midrule
\multicolumn{2}{l}{\textbf{Llama3.2-1B}} & 52.26 &  &  &  &  &  &  & 5.94 \\
 & Compression-Tokens (Causal) &  & 41.97 & 40.07 & 35.08 & 35.24 & 22.50 & 23.14 &  \\
 & Compression-Tokens (Bidirectional) &  & 44.58 & 44.02 & 36.36 & 38.77 & 23.84 & 23.88 &  \\
 & Mean-Pooling &  & 49.90 & 46.92 & 33.84 & 39.59 & 17.67 & 21.17 &  \\
\midrule
\bottomrule
\end{tabular}
    }
    \caption{ParaphraseRC $F_1$.}
    \label{tab:primary_results_paraphrase_rc}
\end{table*}

\section{LLM Usage}\label{app:llm}
LLMs (specifically, ChatGPT and Claude) were used in the process of writing this paper for creating tables and figures, rephrasing, and proof-reading.

\end{document}